\documentclass[10pt,journal,compsoc]{IEEEtran}
\ifCLASSOPTIONcompsoc
  \usepackage[nocompress]{cite}
\else
  \usepackage{cite}
\fi
\ifCLASSINFOpdf
   \usepackage[pdftex]{graphicx}
\else
   \usepackage[dvips]{graphicx}
\fi
\usepackage{array}
\usepackage{fixltx2e}

\usepackage{stfloats}
\usepackage{url}

\usepackage{bm}
\usepackage{subfigure}
\usepackage{amsmath,amssymb,amsthm}
\usepackage{multirow,array}
\usepackage{color}
\usepackage{algorithmic,algorithm}
\usepackage{lipsum}
\usepackage{booktabs}
\usepackage{graphicx}
\newtheorem{theorem}{Theorem}
\DeclareMathOperator{\Tr}{Tr}
\usepackage{diagbox}
\usepackage{arydshln}
\usepackage{bbding}
\usepackage[colorlinks,linkcolor=blue]{hyperref}

\begin{document}
%
\title{Defensive Few-shot Learning}
%
%
%
%

\author{Wenbin~Li,
        Lei~Wang*,~\IEEEmembership{Senior Member,~IEEE,}
        Xingxing~Zhang,
        Lei~Qi,
        Jing~Huo,
        Yang~Gao*,
        and~Jiebo~Luo,~\IEEEmembership{Fellow,~IEEE}
\IEEEcompsocitemizethanks{\IEEEcompsocthanksitem Wenbin Li, Jing Huo and Yang Gao are with the State Key Laboratory for Novel Software Technology, Nanjing University, China, 210023 (e-mail: liwenbin@nju.edu.cn; huojing@nju.edu.cn; gaoy@nju.edu.cn).
\IEEEcompsocthanksitem Lei Wang is with the School of Computing and Information Technology, University of Wollongong, Australia (e-mail: leiw@uow.edu.au).
\IEEEcompsocthanksitem Xingxing Zhang is with the Department of Computer Science, Tsinghua University, China (e-mail: xxzhang2020@mail.tsinghua.edu.cn).
\IEEEcompsocthanksitem Lei Qi is with the School of Computer Science and Engineering, Southeast University, China (e-mail: qilei@seu.edu.cn).
\IEEEcompsocthanksitem Jiebo Luo is with the Department of Computer Science, University of Rochester, America (e-mail: jluo@cs.rochester.edu).
\IEEEcompsocthanksitem *Corresponding authors: Lei Wang; Yang Gao.
    }}
\IEEEtitleabstractindextext{%
\begin{abstract}
This paper investigates a new challenging problem called {\em defensive few-shot learning} in order to learn a robust few-shot model against adversarial attacks. Simply applying the existing adversarial defense methods to few-shot learning cannot effectively solve this problem. This is because the commonly assumed sample-level distribution consistency between the training and test sets can no longer be met in the few-shot setting. To address this situation, we develop a general \textit{defensive few-shot learning (DFSL) framework} to answer the following two key questions: (1) how to transfer adversarial defense knowledge from one sample distribution to another? (2) how to narrow the distribution gap between clean and adversarial examples under the few-shot setting? To answer the first question, we propose an \emph{episode-based adversarial training mechanism} by assuming a \textit{task-level distribution consistency} to better transfer the adversarial defense knowledge. As for the second question, within each few-shot task, we design two kinds of distribution consistency criteria to narrow the distribution gap between clean and adversarial examples from the \textit{feature-wise} and \textit{prediction-wise} perspectives, respectively. Extensive experiments demonstrate that the proposed framework can effectively make the existing few-shot models robust against adversarial attacks. Code is available at \UrlFont{https://github.com/WenbinLee/DefensiveFSL.git}.
\end{abstract}

}

\maketitle

\IEEEdisplaynontitleabstractindextext

%
\IEEEpeerreviewmaketitle

\IEEEraisesectionheading{\section{Introduction}\label{sec:introduction}}

%
%
%
%
\IEEEPARstart{D}{eep} convolutional neural networks (CNNs)~\cite{lecun1989backpropagation,krizhevsky2012imagenet} have obtained impressive successes on a variety of computer vision tasks especially in image classification~\cite{he2016deep,gu2018recent}. Unfortunately, several pieces of recent work~\cite{goodfellow2014explaining,AADsurvey2018} have shown that these CNN models are vulnerable to adversarial examples (attacks), which are crafted based on original clean examples (\textit{i.e.}, images) with imperceptible perturbations. It means that the CNN models could easily misclassify these adversarial examples. Therefore, how to learn robust CNN models that can effectively defend against adversarial attacks becomes a crucial problem. Recently, many adversarial defense methods, especially adversarial training based methods, have been proposed and considerably improved the robustness of deep CNN models~\cite{ZhengSLG16,madry2017towards,miyato2018virtual,athalye2018obfuscated,kannan2018adversarial,xu2019structured,tsipras2019robustness,karim2020adversarial}.

The existing adversarial training based studies mainly focus on generic image classification and try to make generic deep models robust against adversarial attacks. They basically rely on a large amount of labeled data available for each class. However, the robustness of few-shot learning models against adversarial attacks is rarely considered in the literature. This problem is truly important in many real applications, where we not only face the limitation of only accessing a few labeled samples for new, unseen classes, but also must be concerned about the robustness of the intelligent deep learning systems. For example, face recognition on automated teller machine (ATM) and mobile phones~\cite{ciubotaru2019revisiting} could become vulnerable to adversarial attacks and ends up with serious consequences. Similarly, in other applications such as malware classification~\cite{ale2020few} and medical image analysis~\cite{tang2021recurrent,feng2021interactive}, the corresponding deep learning systems could be compromised by adversarial attacks if the robustness issue is not sufficiently addressed. What's worse, under the few-shot setting, deep models could become more vulnerable to adversarial attacks due to the serious lack of training samples~\cite{schmidt2018adversarially}. Therefore, how to learn a robust few-shot model defensive against adversarial attacks is raised in this work as a new and challenging issue, and we name it \textit{defensive few-shot learning} throughout this paper.

However, we cannot directly apply the existing adversarial defense methods in the way that they are applied to generic image classification, to few-shot learning to effectively tackle the defensive few-shot image classification issue defined in this paper. The reasons are in two folds. First, as proved by recent work~\cite{schmidt2018adversarially}, adversarially robust generalization requires access to more data, but the few-shot setting only has access to significantly fewer training samples (\textit{e.g.,} only $1$ or $5$ samples per class) than generic image classification. This makes defensive few-shot image classification much more difficult to achieve. Second, generic image classification can usually safely assume the sample-level distribution consistency, \textit{i.e.}, the independently and identically distributed (\textit{i.i.d.}) assumption, between the training and test sets. However, this assumption cannot be made anymore in the few-shot case. This is because due to the serious scarcity of training samples, few-shot learning usually has to resort to a large but class-disjoint auxiliary set to learn transferable knowledge. This means that the actual training set often has a somewhat different sample distribution from the test set of the target few-shot task.

Therefore, defensive few-shot learning should be investigated as a new challenging issue, which is different from both generic adversarial training and standard few-shot learning. To address this issue, two key questions need to be answered: (1) how to transfer adversarial defense knowledge from one sample distribution (\textit{i.e.}, the auxiliary set) to another (\textit{i.e.}, the unseen few-shot task)? (2) how to narrow the distribution gap between clean and adversarial examples under the few-shot setting? Note that the goal of this paper is to study how to make existing few-shot learning models defensive against adversarial attacks, rather than design another new few-shot learning method. Therefore, to achieve this goal and find answers to the above two questions, we propose a new and general \textit{defensive few-shot learning (DFSL) framework}, which can be efficiently tailored to the existing few-shot learning methods to learn a defensive few-shot model.

Specifically, to answer the first question, we make a \textit{task-level distribution consistency assumption}, instead of the original \textit{sample-level distribution consistency assumption} used in current adversarial defense methods, between the training set (\textit{i.e.}, the auxiliary set) and test set (\textit{i.e.}, the target few-shot tasks). Based on this assumption, we propose an \textit{episode-based adversarial training (ET) mechanism} to transfer the adversarial defense knowledge between two sample distributions, by adversarially training a defensive few-shot model on thousands of adversary-based few-shot tasks (episodes) constructed from the training set. The core is that, from the sample-level perspective, the distribution between the training set and test set may vary. However, from the task-level perspective (\textit{i.e.,} a higher level), the task distribution between them could be assumed to be similar. According to this assumption, \textit{i.e.,} both the training set and test set share a similar (or the same) task-distribution, the model adversarially trained on the adversary-based few-shot tasks constructed from the training set can generalize well to the similar adversary-based few-shot tasks of the test set. In this way, the adversarial defense knowledge can be transferred from one sample distribution to another through such a task-level assumption. Note that although the transferability of adversarial examples, \textit{i.e.,} adversarial examples can be transferred across different models, has been a common sense in the adversarial learning community~\cite{xie2019improving}, the research on the transferability of defense knowledge has not been well investigated in the literature of adversarial learning. We highlight that we take a small step in this direction.

Furthermore, within each adversary-based few-shot task, we shall enforce a distribution consistency between the clean and adversarial examples (images) like the existing adversarial defense methods to further improve the classification performance, \textit{i.e.,} the second question above. Note that because the existing adversarial defense methods are mainly designed for generic classification problems, which are assumed to be able to access sufficient training examples, these methods mainly work with the pooled global (logit) representations of the clean and adversarial images and aim to make them consistent. However, in DFSL, \textit{i.e.,} in the few-shot setting, we can only have access to a small amount of data, which makes the second question more challenging. To alleviate this scarcity issue of training data under the few-shot setting, we propose to switch to the richer local descriptors instead of the global representations to represent each clean and adversarial image. For each image, we can extract a large number of local descriptors to represent this image. Based on such local-descriptor-based representations, we especially propose a novel kind of \textit{feature-wise consistency criteria} to enforce a local-descriptor-based distribution consistency between the clean and adversarial examples. Specifically, we design two distribution measures, \textit{i.e.,} a \textit{Kullback-Leibler divergence based distribution measure (KLD)} and a \textit{task-conditioned distribution measure (TCD)}, to align the local-descriptor-based distributions between the clean and adversarial examples. In addition, following the existing adversarial defense methods~\cite{kannan2018adversarial,miyato2018virtual,ZhangICML2019}, we can also enforce a kind of \textit{prediction-wise consistency} between each clean example and its adversarial counterpart, by making their predicted posterior probability distributions of the classes to be similar. However, the existing methods usually employ tight regularizers to achieve this goal, which we find is no longer suitable for the few-shot setting in DFSL. The reason is that, in the few-shot setting, the test set generally has a certain distribution gap with respect to the training set. Using such tight regularizers on the training set will weaken the adversarially robust generalization ability of the defensive few-shot models on the test set. To tackle this issue, different from the existing methods, we propose a slacker \textit{Symmetric Kullback-Leibler divergence measure (SKL)} to obtain a better adversarially robust generalization ability.
In summary, by taking the above two aspects (\textit{i.e.,} the task-level distribution consistency assumption and distribution consistency criteria within each task) into consideration, our proposed DFSL framework is able to learn a defensive few-shot model against adversarial attacks.

In addition, we find that the existing adversarial defense methods often report two kinds of classification accuracy, \textit{i.e.}, clean example accuracy and adversarial example accuracy, to show the effectiveness of their proposed defense methods. However, there may be a trade-off between these two kinds of accuracies~\cite{tsipras2019robustness,ZhangICML2019}, which means that the gain of the adversarial accuracy can be the loss of the clean accuracy or vice versa. This makes a direct comparison of different methods awkward, if not impossible. Therefore, it is desirable to have a unified criterion to facilitate the evaluation and comparison of different defense methods under the same principle, which has been largely overlooked in the existing literature. To this end, we propose a unified $\mathcal{F}_{\beta}$ score to conveniently evaluate the overall performance of different defense methods under the same principle.

Last but not least, we also find that \textit{randomness matters} in adversarial training, especially in defensive few-shot learning. In other words, the randomness will make the comparison between different defense methods unfair. This is because different runs of the same defense method on the same platform may end up with quite different results due to random initializations of network parameters, random data shuffles, or the randomness of CUDA and CuDNN backends. In particular, the last point, \textit{i.e.,} the randomness of CUDA and CuDNN backends, is easily overlooked (See Section~\ref{Section_5_6} for more details). Therefore, in this work, to make the comparison of different defense methods fairer and make the results reproducible, we completely control the randomness by fixing both the network initialization and data shuffle, including the randomness of both CUDA and CuDNN backends, for all comparison methods. More importantly, all the comparison methods are implemented under the same framework with the same single codebase.

In summary, the main contributions of this paper are:
\begin{itemize}
    \item We define a new challenging issue, \textit{i.e.}, \textit{defensive few-shot learning (DFSL)}, for the first time in the literature. This poses two key questions (challenges): how to transfer defense knowledge during adversarial training and narrow the distribution gap between clean and adversarial examples under the few-shot setting.
    \item We propose a novel and general DFSL framework to address the above two challenges, by performing an \textit{episode-based adversarial training} at the task level and enforcing the distribution consistency between clean and adversarial examples from the \textit{feature-wise} or \textit{prediction-wise} perspectives within each task.
    \item We further design three new distribution consistency criteria, \textit{i.e.,} \textit{Kullback-Leibler divergence based distribution measure (KLD)}, \textit{task-conditioned distribution measure (TCD)}, and \textit{Symmetric Kullback-Leibler divergence measure (SKL)}, to specially narrow the distribution gap under the defensive few-shot setting.
    \item We tailor the proposed DFSL framework to the state-of-the-art few-shot learning methods and conduct extensive experiments on six benchmark datasets to verify the effectiveness of this framework. This provides rich baseline results for this new problem, \textit{i.e.}, defensive few-shot learning, and meanwhile, facilitates future research on this topic.
\end{itemize}

\section{Related Work}
Our work is related to few-shot learning and adversarial training, both of which have a large body of work. Here, we only discuss the most relevant studies in these two fields. In addition, we will introduce the episodic training mechanism used in the standard few-shot learning and review multiple state-of-the-art attack methods.

\textbf{Few-shot learning (FSL)} attempts to learn a classifier with good generalization capacity for new unseen classes with only a few samples~\cite{fei2006one,VinyalsBLKW16,allen2019infinite,li2019DN4,jamal2019task,ChenFCJ_AAAI19,ChenFW00H_CVPR2019,ye2020heterogeneous,li2020ADM,dong2021learning,lu2020learning,sun2020meta,KimKK20a_ECCV2020,ReNet_CVPR_2021,li2021libfewshot}. Due to the scarcity of data, a large-scale but class-disjoint auxiliary set is generally used to learn transferable knowledge for the target few-shot tasks. Specifically, in~\cite{VinyalsBLKW16}, Vinyals \emph{et al.} propose a Matching Net by directly comparing the query images with the support classes. In particular, in the work of Matching Net, they also propose an episodic training mechanism, which is widely adopted and taken as the default in the subsequent studies. Along this way, a variety of methods have been proposed, such as ProtoNet~\cite{snell2017prototypical}, RelationNet~\cite{sung2018learning}, IMP~\cite{allen2019infinite}, CovaMNet~\cite{li2019CovaMNet}, CAN~\cite{CAN_NeurIPS2019}, DeepEMD~\cite{deepemd_CVPR2020} and DN4~\cite{li2019DN4}.

As a special problem setting in FSL, the proposed \textit{defensive few-shot learning (DFSL)} aims to make the existing FSL methods robust against adversarial attacks.

\textbf{Episodic training mechanism} plays an important role in the above FSL methods, which tries to train a few-shot model by constructing tens of thousands of simulated episodes (tasks) from an auxiliary set. To be specific, each episode (task) is a simulation of the target few-shot task, which also consists of two akin subsets, \textit{i.e.}, a support set and a query set. At each iteration, one episode (task) is adopted to train the current model.

However, although the promising performance of the episodic training mechanism has been verified in the standard FSL methods~\cite{VinyalsBLKW16,snell2017prototypical,zhang2020iept}, the effectiveness of this mechanism under the defensive few-shot setting has not been investigated. In this work, we interpret this mechanism from the perspective of task-level distribution consistency and develop transferable adversarial defense upon it.

\textbf{Adversarial training (AT)} is a specific training mechanism that trains a model with both adversarial examples and clean examples in order to make the model robust against adversarial attacks~\cite{goodfellow2014explaining,ZhengSLG16,madry2017towards,na2018cascade,tsipras2019robustness,ZhangICML2019}. For example, to improve the robustness of semi-supervised classification, Miyato \emph{et al.}~\cite{miyato2018virtual} propose a semi-supervised virtual adversarial training method (VAT) by calculating the KL divergence between the predictions on the clean examples and the adversarial examples. Similarly, Kannan \emph{et al.}~\cite{kannan2018adversarial} present an adversarial logit pairing (ALP) strategy, encouraging similar logit representations (\textit{i.e.,} unscaled probability distributions) of the clean and the corresponding adversarial examples. Recently, Song \emph{et al.}~\cite{MMDICLR2019} introduce domain adaptation into adversarial training (ATDA) to learn domain invariant representations for both clean and adversarial domains. Zhang \emph{et al.}~\cite{ZhangICML2019} theoretically identify a trade-off between robustness and accuracy, and propose TRADES to optimize a regularized surrogate loss.

The main differences between our DFSL framework and these methods are: (1) the above methods only consider the generic image classification setting, rather than the more challenging few-shot setting considered in this paper; (2) these methods mainly focus on the global prediction-wise consistency between the clean and adversarial examples, while our DFSL framework proposes a new feature-wise consistency from a perspective of local-descriptor-based distribution consistency, which provides an effective way for capturing the distributions of both clean and adversarial examples in the few-shot case; (3) all these above methods, taken as regularizers, can be tailored into the proposed DFSL framework.

\textbf{Attack Methods} can not only be used to attack or test a system, but can also be employed to make this system more robust against such kinds of attacks. A variety of attack methods have been proposed in the literature~\cite{LBFGS2013, goodfellow2014explaining,kurakin2016adversarial,CW2017,madry2017towards,miyato2018virtual}. For example, both L-BFGS~\cite{LBFGS2013} and C\&W~\cite{CW2017} attempt to find an adversarial example through an optimization way, by optimizing a constrained minimization problem, \textit{i.e.,} a minimum $\ell_2$ norm distance between this misclassified adversarial example and the corresponding clean example. These kinds of methods are generally time-consuming due to the optimization process adopted. Differently, Goodfellow \emph{et al.}~\cite{goodfellow2014explaining} propose a one-step \textit{fast gradient sign method} (FGSM), which generates an adversarial example through a single backward propagation of the neural network with respect to the clean input. In this way, the adversarial examples can be quickly constructed because FGSM does not need the optimization process. After that, a stronger iterative variant of FGSM, \textit{i.e.}, \textit{projected gradient descent} (PGD) is proposed in~\cite{madry2017towards}, which mainly applies the FGSM iteratively for multiple times with a small step size. Different from L-BFGS and C\&W, which try to optimize an $\ell_2$ norm distance, both FGSM and PGD are usually optimized for an $\ell_\infty$ norm distance metric. Because FGSM is much faster than other attack methods to generate an adversarial example, we will mainly adopt it to generate adversarial examples in the training phase and adopt it to test the robustness of the learned DFSL models. In addition, PGD is also employed and evaluated in Section~\ref{Section_5_8}.

\begin{figure*}[!tp]
\centering
\includegraphics[width=0.55\textwidth]{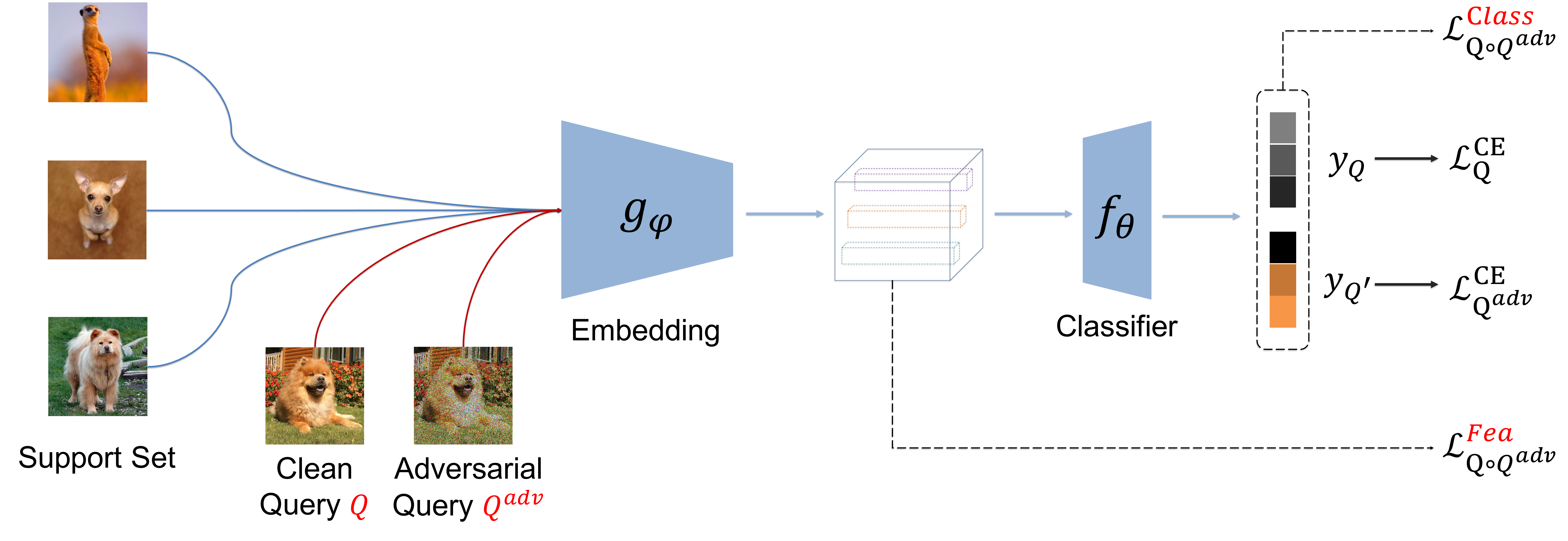}
\vspace{-3mm}
\caption{Proposed framework of the \textit{defensive few-shot learning (DFSL)} on a $3$-way $1$-shot task, which is decoupled into two modules, \textit{i.e.,} a feature embedding module $g_{\varphi}$ and a classifier module $f_{\theta}$. Specifically, a support set $\mathcal{S}$, a clean query set $\mathcal{Q}$ and an adversarial query counterpart $\mathcal{Q}^{adv}$ are fed into the model, supervised by three kinds of losses, \textit{e.g.,} the cross entropy loss (\textit{i.e.,} $\mathcal{L}^\text{CE}_Q$ and $\mathcal{L}^\text{CE}_{Q^\text{adv}})$, the feature-wise loss $\mathcal{L}^\text{Fea}_{Q\circ Q^\text{adv}}$ and the prediction-wise loss $\mathcal{L}^\text{Class}_{Q\circ Q^\text{adv}}$, which will become clear shortly.}
\label{fig_framework}
\end{figure*}

\section{The Proposed Defensive Few-shot Learning Framework}
In this section, we first introduce the notations used in this paper, and then present the definition of the new topic, \textit{defensive few-shot learning (DFSL)}. Finally, we provide the details of the proposed defense framework.

\subsection{Notation}
\label{notation}
Following the literature, let $\mathcal{S}$ and $\mathcal{Q}$ denote the support set and query set in an FSL task, which corresponds to the training set and test set in generic image classification, respectively. Differently, $\mathcal{S}$ contains $C$ classes but only has $K$ images per class (\textit{e.g.}, $K=1$ or $K=5$). $\mathcal{A}$ indicates an additional auxiliary set $\mathcal{A}$, which contains a larger number of classes and samples than $\mathcal{S}$ but has a totally disjoint label space with $\mathcal{S}$.

Let $g_\varphi(\cdot)$ denote a convolutional neural network based embedding module, which can learn feature representations for any input image $\bm{x}$, \textit{i.e.,} $g_\varphi(\bm{x})$. Suppose $f_\theta(g_\varphi(\bm{x}), \mathcal{S})$ be a classifier module, which assigns a class label $y$ for a query image $\bm{x}$ in $\mathcal{Q}$, according to $\mathcal{S}$. Note that the classifier module $f_\theta(\cdot)$ can be integrated with the embedding module $g_\varphi(\cdot)$ into a unified network and trained in an end-to-end manner. The cost function $\mathcal{L}\big(f_\theta(g_\varphi(\bm{x}), \mathcal{S}), y\big)$ is denoted as $\mathcal{L}(\bm{x},\mathcal{S},y)$ for simplicity.

It is easy to use a small perturbation $\bm{\delta}$ to construct an adversarial image $\bm{x}^\text{adv}\!=\!\bm{x}\!+\!\bm{\delta}$ to fool the classifier, making $f_\theta(g_\varphi(\bm{x}^\text{adv}), \mathcal{S})\!\ne\!y$. Generally, the clean image $\bm{x}$ and the adversarial image $\bm{x}^\text{adv}$ are perceptually indistinguishable, and their difference (\textit{i.e.,} the perturbation $\bm{\delta}$) can be bounded by a distance metric $D(\bm{x},\bm{x}^\text{adv})\le\epsilon$, such as the $\ell_\infty$ norm. That is to say, if $\epsilon$ indicates the maximum magnitude of the perturbation $\bm{\delta}$, we have $\|\bm{\delta}\|_\infty\le\epsilon$. Note that, all clean images are normalized into a range of $[0,1]$, and all adversarial images are clipped into the same range. Following the work in~\cite{kurakin2016adversarial}, a white-box attack setting~\cite{AADsurvey2018} is employed to generate all the training adversarial images in this paper.

\subsection{Defensive Few-shot Learning (DFSL)}
As mentioned above, a few-shot task normally consists of a support set $\mathcal{S}$ and a query set $\mathcal{Q}$. Given $\mathcal{S}$, which has $C$ classes with $K$ images per class, the target of FSL is to infer the correct class label for each unlabeled sample in $\mathcal{Q}$. This setting is typically called a $C$-way $K$-shot classification setting. Since the number of $K$ is generally small (\textit{e.g.}, $1$ or $5$), it is almost impossible to learn an effective classifier only from $\mathcal{S}$. Therefore, in FSL, an additional auxiliary set $\mathcal{A}$ is usually adopted to learn transferable knowledge to help the classification on $\mathcal{Q}$.

Unlike the standard FSL, here, we mainly focus on how to learn a robust few-shot classification model to defend against adversarial attacks, \textit{i.e.}, \textit{defensive few-shot learning (DFSL)}. In DFSL, we always assume that the adversary is capable of manipulating the query images in the query set $\mathcal{Q}$, but doesn't have access to the support set $\mathcal{S}$. In other words, for one few-shot task which has been adversarially attacked by the adversary, we will have three kinds of sets in this task, \textit{i.e.}, a support set $\mathcal{S}$, a clean query set $\mathcal{Q}$, and an adversarial query set $\mathcal{Q}^\text{adv}$. Without loss of generality, we assume each image in $\mathcal{Q}$ has one corresponding adversarial counterpart in $\mathcal{Q}^\text{adv}$. Typically, we call an attacked few-shot task as an \textit{adversary-based few-shot task}. Our goal in DFSL is to learn a robust model which can correctly classify query images no matter if they are manipulated, \textit{i.e.}, query images in both $\mathcal{Q}$ and $\mathcal{Q}^\text{adv}$. The framework can be seen in Fig.~\ref{fig_framework}.

\subsection{Analysis and Answers to the Two Questions}
\label{analysis_task}
Due to the scarcity of training data, few-shot models need to learn transferable knowledge from a class-disjoint auxiliary set, which generally has a somewhat different sample distribution with respect to the target few-shot task. This situation makes the problem of DFSL quite different from the generic adversarial defense problems, and thus raises two new questions on DFSL: (1) how to transfer adversarial defense knowledge between two sample distributions (\textit{i.e.}, $\mathcal{A}$ and $\mathcal{S}$)? (2) how to narrow the distribution gap between the clean and adversarial examples in a specific adversary-based few-shot task (\textit{i.e.}, the distribution gap between $\mathcal{Q}$ and $\mathcal{Q}^\text{adv}$)?

\textbf{The first question} does not exist in generic image classification, because we can usually make a distribution consistency assumption between the training and test sets (\textit{i.e.}, independently and identically distributed data) to guarantee the model trained on the training set can generalize well to the test set. Such an assumption is also implicitly assumed in the existing adversarial defense methods~\cite{goodfellow2014explaining,kannan2018adversarial,MMDICLR2019}. However, in DFSL, the auxiliary set (for training) $\mathcal{A}$ has a totally class-disjoint label space with the support set $\mathcal{S}$ in the target adversary-based few-shot task. Since the sample distribution of $\mathcal{A}$ is relatively different from the sample distribution of $\mathcal{S}$, the generalization performance on the target data set (\textit{i.e.}, $\mathcal{S}$) cannot be well guaranteed. As a result, the adversarial defense knowledge learned on $\mathcal{A}$ by directly using the existing adversarial defense methods is hard to be transferred to $\mathcal{S}$ in the target task. This is why we cannot directly employ the existing adversarial defense methods to address the DFSL problem. This will be demonstrated in the experimental part later.

\begin{figure*}[!tp]
\centering
\includegraphics[width=0.65\textwidth]{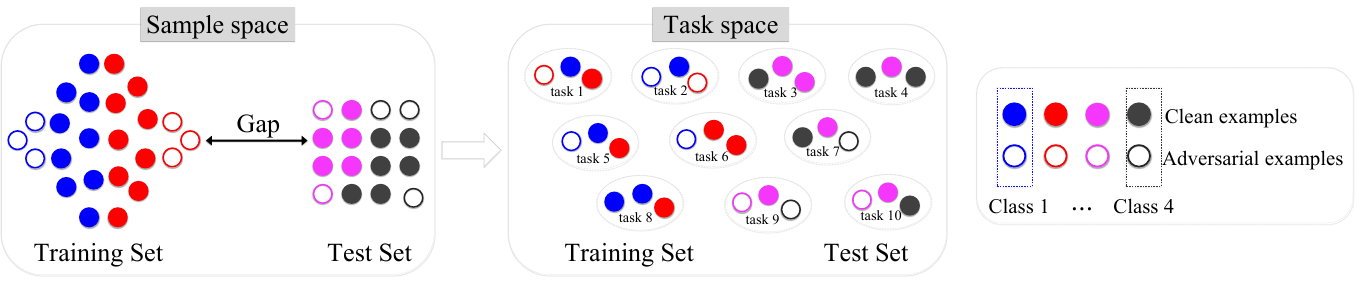}
\vspace{-3mm}
\caption{Changing the distribution consistency assumption from the sample-level to the task-level. As seen, the sample distribution gap (difference) between the training and test sets in the sample space may be significant, while the task distribution gap (difference) in the task space between them could be small because of the task similarity. Each color indicates one class. Solid and hollow circles indicate the clean and adversarial examples, respectively.}
\label{fig_distribution}
\end{figure*}

The above phenomenon can be visualized as the left side in Fig.~\ref{fig_distribution}, \textit{i.e.}, there may be a large distribution gap between the training set (\textit{i.e.}, $\mathcal{A}$) and test set (\textit{i.e.}, $\mathcal{S}$) in the sample space, making the adversarial defense knowledge hard to transfer. To address this issue, inspired by the episodic training mechanism~\cite{VinyalsBLKW16}, we can assume the distribution consistency in a task space instead of the sample space (see the right side of Fig.~\ref{fig_distribution}). From the perspective of the task-level distribution consistency between the training set and test set, we can construct a large number of adversary-based few-shot tasks within the auxiliary set $\mathcal{A}$, by simulating the target adversary-based few-shot task in the test set. In doing so, the sample distribution gap can be dealt with by leveraging the task similarity across the training and test sets. In other words, although two sample distributions may be relatively different from the lower sample-level consistency perspective, they can still have similarities from a higher task-level consistency perspective.

The contribution of our work here is to attempt to develop transferable adversarial defense upon the episodic training mechanism~\cite{VinyalsBLKW16} in the new setting of DFSL and especially present a new \textit{episode-based adversarial training mechanism} for the DFSL problem. To the best of our knowledge, addressing this new problem in such a setting is the first time in the literature.

\textbf{The second question} is essentially a common issue that the existing adversarial defense methods are working on. According to the adversarial learning literature~\cite{MMDICLR2019}, there is usually a large distribution gap between the clean and adversarial examples, making the deep models prone to misclassifying the manipulated adversarial examples. Therefore, the latest adversarial defense methods~\cite{kannan2018adversarial,MMDICLR2019,ZhangICML2019} are focusing on how to narrow such a distribution gap between the clean and adversarial examples to improve the classification performance. However, these methods are mainly designed for the generic classification problems, rather than the more challenging few-shot setting considered in this paper. In fact, few-shot models are more vulnerable to the adversarial examples than generic deep models, because of the serious scarcity of labeled training samples in each class. The evidence can be seen in a recent work of adversarial training~\cite{schmidt2018adversarially}, where it is proved that adversarially robust generalization requires access to more data. As a result, how to make few-shot models robust against adversarial attacks, especially how to narrow the distribution gap between the clean and adversarial examples under the few-shot setting, is of significance but challenging.

Multiple adversarial defense methods have been proposed to address the above issue (\textit{i.e.}, the distribution gap between clean and adversarial examples) in the generic classification problems. They mainly focus on using global representations for clean and adversarial images. For example, ATDA~\cite{MMDICLR2019} tries to minimize the distribution shift (gap) between the clean and adversarial examples by minimizing the covariance distance between their covariance matrices, which are estimated upon the global logit representations. Clearly, ATDA can not be directly applied in DFSL because of the scarcity of training examples, which makes the covariance matrix hard to be reliably estimated. Also, these existing adversarial defense methods are developed under an implicit \textit{i.i.d.} assumption for the training set and test set. In this case, they usually can safely apply some tight regularization criteria, \textit{e.g.,} the $\ell_2$ regularization loss presented in ALP~\cite{kannan2018adversarial}. However, it is found in this work that such tight regularization criteria will somewhat weaken the generalization ability of the few-shot model in DFSL, especially when the test set has a large distribution gap with the training set.

Therefore, different from the existing adversarial defense methods, we propose a new kind of \textit{feature-wise consistency criteria} upon richer local descriptors to narrow the distribution gap between the clean and adversarial examples in each adversary-based few-shot task. The reason is that, even in the few-shot setting, a large number of local descriptors can still be extracted from each image. This can be regarded as a kind of natural data augmentation, which can significantly enrich the amount of data. On the other hand, the local descriptors generally contain abundant subtle information on the visual content. The existing adversarial defense methods usually work on the global logit representations that are built on the pooled local descriptors. This could cause the loss of information, and the impact could become pronounced in the case of a few-shot setting. Therefore, by using a kind of local-descriptor-based feature-wise consistency criteria, the scarcity problem of training examples in DFSL can be well alleviated.

In addition, we design a slacker \textit{prediction-wise consistency criterion} to address the issue caused by existing tight regularization criteria to enhance the adversarially robust generalization ability of the DFSL models. This is because there is usually a distribution gap between the training set and test set in DFSL, making the existing tight regularization criteria easily suffer from the overfitting problem. In contrast, by using a slack criterion, the DFSL models can gain more generalization performance.

\subsection{The Proposed Adversarial Defense Methods}
The above analysis motivates us to propose a novel framework for DFSL in Fig.~\ref{fig_framework}, where a defensive few-shot model $\mathcal{F}$ is explicitly decoupled into an embedding modules $g_\varphi$ and a classifier module $f_\theta$, \textit{i.e.,} $\mathcal{F}=g_\varphi\circ f_\theta$. To learn a robust few-shot model $\mathcal{F}$, we first employ an \textit{episode-based adversarial training (ET) mechanism} to transfer the adversarial defense knowledge at the task level. Next, we further design two kinds of \textit{distribution consistency criteria} associated with the two modules (\textit{i.e.,} $g_\varphi$ and $f_\theta$) in $\mathcal{F}$, respectively. The former is to address the above first question, and the latter is used to tackle the second question above.

\subsubsection{Episode-based Adversarial Training (ET)}
\label{Section_3_4_1}
According to the analysis of the first question in Section~\ref{analysis_task}, we propose the following \textit{episode-based adversarial training (ET) mechanism}. Specifically, let $\{\langle\mathcal{S}_1, \mathcal{Q}_1, \mathcal{Q}_1^\text{adv}\rangle,\ldots, \langle\mathcal{S}_n, \mathcal{Q}_n, \mathcal{Q}_n^\text{adv}\rangle\}$ be a set of adversary-based few-shot tasks randomly constructed from the auxiliary set $\mathcal{A}$. The objective function of our ET mechanism can be formulated as,
\begin{equation}\label{fun1}\small
\begin{split}
 \mathnormal{\Gamma}\!=\!\underset{\mathcal{\varphi,\theta}}{\arg\min}\sum_{i=1}^n\sum_{\bm{x}\in \mathcal{Q}_i}\Big(\mathcal{L}^\text{CE}(\bm{x},\mathcal{S}_i,y)\!+\!
 \underset{\bm{x}^\text{adv}\!=\!\bm{x}\!+\!\bm{\delta}}{\max}\mathcal{L}^\text{CE}(\bm{x}^\text{adv},\mathcal{S}_i,y)\Big)\,,
\end{split}
\end{equation}
where $\varphi$ and $\theta$ denote the parameters of the embedding module $g_\varphi(\cdot)$ and the classifier module $f_\theta(\cdot)$ in a DFSL model, respectively, and $\mathcal{L}^\text{CE}(\cdot)$ denotes the cross-entropy loss (\textit{i.e.,} $\mathcal{L}^\text{CE}_Q$ and $\mathcal{L}^\text{CE}_{Q^\text{adv}}$ in Fig.~\ref{fig_framework}). Other notations have been defined in Section~\ref{notation}. The core idea here is to simulate the target adversary-based few-shot task by conducting a lot of similar adversary-based few-shot tasks with the auxiliary set $\mathcal{A}$. In this way, we can build a task-based distribution to deal with the gap among different sample distributions.

At each training step, we generate adversarial examples (\textit{i.e.}, $\mathcal{Q}_i^\text{adv}$) within each of the sampled few-shot tasks (\textit{i.e.}, $\langle\mathcal{S}_i, \mathcal{Q}_i\rangle$) based on the current model, and meanwhile, inject these adversarial examples into the current few-shot tasks (\textit{i.e.}, $\langle\mathcal{S}_i, \mathcal{Q}_i, \mathcal{Q}_i^\text{adv}\rangle$). Both clean and adversarial examples (\textit{i.e.}, $\mathcal{Q}_i$ and $\mathcal{Q}_i^\text{adv}$) are used to train the model, enhancing its capability to defend adversarial attacks.

Essentially, the \textit{episode-based adversarial training (ET) mechanism} can be regarded as an integration of \textit{episodic training mechanism (Ep)}~\cite{VinyalsBLKW16} and \textit{adversarial training (AT)}~\cite{goodfellow2014explaining}, \textit{i.e., ET=Ep+AT}. Seemingly, the integration of Ep and AT for addressing the DFSL problem is straightforward. However, as mentioned in Section~\ref{analysis_task}, the key contribution here is that we verify that ET can indeed effectively transfer the adversarial defense knowledge between different sample distributions, which has not been specifically investigated or confirmed in the literature. More importantly, we further examine this ET mechanism and improve it to consider the distribution gap between the clean and adversarial examples within each adversary-based few-shot task. To tackle this limitation, especially in the few-shot setting, we develop new criteria upon this ET mechanism to enforce the distribution consistency between the clean and adversarial examples to further boost the classification performance.

\subsubsection{Feature-wise Consistency Criteria}
\label{Section_3_4_2}
Based on the analysis of the second question in Section~\ref{analysis_task}, we shall further enforce a distribution consistency between the clean example and its adversarial counterpart, within each adversary-based few-shot task. Specifically, we can enforce this consistency by making their feature representations (or distributions) similar from a \textit{feature-wise} perspective. As discussed in Section~\ref{analysis_task}, instead of the global features, we adopt the much richer local descriptors to represent each image, and design two distribution measures (regularizers), \textit{i.e.,} a \textit{Kullback-Leibler divergence based distribution measure (KLD)} and a \textit{task-conditioned distribution measure (TCD)}, built on the local-descriptor-based distributions of both clean and adversarial examples. Moreover, inspired by ATDA~\cite{MMDICLR2019}, we also implement a new \textit{local-descriptor-based unsupervised domain adaptation measure (Local-UDA)}.

Given a specific adversary-based few-shot task $\langle\mathcal{S}, \mathcal{Q},\mathcal{Q}^\text{adv}\rangle$, $\mathcal{S}=\{\bm{s}_1, \bm{s}_2,\ldots, \bm{s}_N\}$ denotes the support set of a $C$-way $K$-shot few-shot task (where $N=C\times K$), $\mathcal{Q}=\{\bm{x}_1, \bm{x}_2,\ldots, \bm{x}_{N_q}\}$ indicates the query set which has $N_q$ query images, and $\mathcal{Q}^\text{adv}=\{\bm{x}_1^\text{adv}, \bm{x}_2^\text{adv},\ldots, \bm{x}_{N_q}^\text{adv}\}$ is the corresponding adversarial query set. As seen in Fig.~\ref{fig_framework}, in DFSL, the embedding module $g_\varphi(\cdot)$ will represent each image $\bm{x}_i$ as a $c\times h \times w$ feature map, \textit{i.e.,} a set of local descriptors $g_\varphi(\bm{x}_i)\!=\![\bm{z}_1,\ldots,\bm{z}_m]\in\mathbb{R}^{c \times m}$ (where $m=h\times w$), instead of a pooled global feature vector. Similarly, for an adversarial example $\bm{x}^\text{adv}$, $g_\varphi(\bm{x}_i^\text{adv})\!=\![\bm{z}_1^\text{adv},\ldots,\bm{z}_m^\text{adv}]\in\mathbb{R}^{c \times m}$. In this work, we assume that the local descriptors of each clean example and each adversarial example follow multivariate normal distributions, \textit{i.e.}, $\bm{z}\!\thicksim\!\mathcal{N}(\bm{\mu}, \bm{\Sigma})$ and $\bm{z}^\text{adv}\!\thicksim\!\mathcal{N}(\bm{\mu}^\text{adv}, \bm{\Sigma}^\text{adv})$. 
$\bm{\mu}$ and $\bm{\Sigma}$ are the mean vector and covariance matrix of $g_\varphi(\bm{x})$, respectively. Also, $\bm{\mu}^\text{adv}$ and $\bm{\Sigma}^\text{adv}$ are the mean vector and covariance matrix of $g_\varphi(\bm{x}^\text{adv})$, respectively.

\textbf{Kullback-Leibler divergence based distribution measure (KLD)}. Since both the clean query image $\bm{x}$ and the adversarial query image $\bm{x}^\text{adv}$ have been represented by a local-descriptor-based distribution, the KLD criterion between $\bm{x}$ and $\bm{x}^\text{adv}$ can be formulated as below,
\begin{equation}\label{fun3}\small
\begin{split}
    \mathcal{L}_\text{KLD}^\text{fea}(\bm{x}, \bm{x}^\text{adv})  &= \frac{1}{2}\Big[\Tr\big((\bm{\Sigma}^\text{adv})^{-1}\bm{\Sigma}\big)+\ln\big( \frac{\det\bm{\Sigma}^\text{adv}}{\det\bm{\Sigma}}\big)  \\
    &+ (\bm{\mu}^\text{adv}-\bm{\mu})^\top(\bm{\Sigma}^\text{adv})^{-1}(\bm{\mu}^\text{adv}-\bm{\mu})-c \Big]\,,
\end{split}
\end{equation}
where $\Tr(\cdot)$ is the trace operation of matrix, $\ln(\cdot)$ denotes the natural logarithm, $\det$ indicates the determinant of a square matrix, and $c$ is the feature dimension of each local descriptor. The purpose of the KLD criterion is to align the local-descriptor-based distributions of $\bm{x}^\text{adv}$ and $\bm{x}$ within each adversary-based few-shot task. The advantage is that KLD can not only narrow the distribution gap between the clean and adversarial examples to some extent, but also enjoys the characteristic of being a slacker criterion which is good for the generalization performance. The reason why KLD is regarded to be slacker is that we do not strictly make the features of two images (\textit{i.e.,} $\bm{x}$ and $\bm{x}^\text{adv}$) exactly the same, but just enforce their local-descriptor-based distributions to be aligned.

\textbf{Task-conditioned distribution measure (TCD)}.
Since we attempt to apply the proposed \textit{episode-based adversarial training (ET) mechanism} to DFSL, many simulated adversary-based few-shot tasks from the auxiliary set will be used to train a robust few-shot model. However, different adversary-based few-shot tasks have their own characteristics. To take the task characteristic into account and make the few-shot model better handle different adversary-based few-shot tasks, we further design a \textit{task-conditioned distribution measure (TCD)} as follows, 
\begin{equation}\label{fun4}\small
\begin{split}
 \mathcal{L}_\text{TCD}^\text{fea}(\bm{x}, \bm{x}^\text{adv}) &=\big(\bm{\mu}-\bm{\mu}^\text{adv}\big)^\top\bm{\Sigma}_{\mathcal{S}}^{-1}\big(\bm{\mu}-\bm{\mu}^\text{adv}\big)\\
 &+\|\bm{\Sigma}_{\mathcal{S}}^{-\frac{1}{2}}\bm{\Sigma}\bm{\Sigma}_{\mathcal{S}}^{-\frac{1}{2}}-\bm{\Sigma}_{\mathcal{S}}^{-\frac{1}{2}}\bm{\Sigma}^\text{adv}\bm{\Sigma}_{\mathcal{S}}^{-\frac{1}{2}}\|_F^2\,,
\end{split}
\end{equation}
where $\bm{\Sigma}_{\mathcal{S}}^{-1}$ denotes the inverse covariance matrix of the support set $\mathcal{S}$. Specifically, we use the local descriptors of all the samples in $\mathcal{S}$ to estimate $\bm{\Sigma}_{\mathcal{S}}$, which can be seen as an overall characterization of the current adversary-based few-shot task. In particular, the first term of Eq.~(\ref{fun4}) is a squared Mahalanobis distance between $\bm{\mu}$ and $\bm{\mu}^\text{adv}$, depending on $\mathcal{S}$. The second term aims to measure the distribution distance between the clean and adversarial examples with the second-order information, which is also depended on $\mathcal{S}$. Note that $\bm{\Sigma}_{\mathcal{S}}^{-\frac{1}{2}}$ is the square root of $\bm{\Sigma}_{\mathcal{S}}^{-1}$, which can be regarded as a transformation matrix to project both clean and adversarial examples of the query set (\textit{i.e.}, $\bm{x}\in\mathcal{Q}$ and $\bm{x}^\text{adv}\in\mathcal{Q}^\text{adv}$) into another new feature space. In the new feature space, we use an approximate $2$-Wasserstein distance~\cite{olkin1982distance} to calculate a distance between the local-descriptor-based distributions of $\bm{x}$ and $\bm{x}^\text{adv}$ (\textit{see the supplementary material for more analyses}). In this way, the distribution gap between the clean and adversarial examples can be adaptively narrowed according to the task information of the current adversary-based few-shot task.

It is worth mentioning that the calculation of matrix square rooting (\textit{e.g.}, obtain $\bm{\Sigma}_{\mathcal{S}}^{-\frac{1}{2}}$ by eigen-decomposition) may make the gradient computation of Eq.~(\ref{fun4}) complicated in backpropagation. To handle this, we propose Theorem~\ref{theorem1} (\textit{the proof is provided in the supplementary material}) to convert Eq.~(\ref{fun4}) into an efficient form without matrix square rooting to facilitate the gradient computation.
\begin{theorem}\label{theorem1}
Suppose $\bm{\Sigma}_1$, $\bm{\Sigma}_2$ and $\bm{\Sigma}$ are all positive semi-definite matrices, and $\bm{\Sigma}^{-\frac{1}{2}}$ is the square root of the inverse of $\bm{\Sigma}$, it can be obtained that 
\begin{equation}\label{FunT1}\small
\begin{split}
 \Tr\big(\|&\bm{\Sigma}^{-\frac{1}{2}}\bm{\Sigma}_1\bm{\Sigma}^{-\frac{1}{2}}-\bm{\Sigma}^{-\frac{1}{2}}\bm{\Sigma}_2\bm{\Sigma}^{-\frac{1}{2}}\|_F^2\big)=
          \Tr\big[\bm{\Sigma}_1\bm{\Sigma}^{-1}\cdot\bm{\Sigma}_1\bm{\Sigma}^{-1}\big] \\
          &-2\Tr\big[\bm{\Sigma}_1\bm{\Sigma}^{-1}\cdot\bm{\Sigma}_2\bm{\Sigma}^{-1}\big] 
          +\Tr\big[\bm{\Sigma}_2\bm{\Sigma}^{-1}\cdot\bm{\Sigma}_2\bm{\Sigma}^{-1}\big]\,.
\end{split}
\end{equation}
\end{theorem}

\textbf{Local-descriptor-based unsupervised domain adaptation measure (Local-UDA).} Seemingly, the recent work of \textit{adversarial training with domain adaptation (ATDA)}~\cite{MMDICLR2019}, especially the unsupervised domain adaptation (UDA) loss it proposed, is closely related to the above proposed KLD or TCD criteria. The difference is clarified as follows. The UDA loss in ATDA~\cite{MMDICLR2019} is originally proposed to minimize the distribution gap of the logit representations (\textit{i.e.,} unscaled probability distributions) between the clean and adversarial examples. That is to say, ATDA is a kind of prediction-wise consistency criteria. Unfortunately, because the original UDA loss in ATDA adopts the logit representations to estimate the mean vectors and covariance matrices of the clean and adversarial examples, it is not suitable anymore for DFSL due to the limited examples under the few-shot setting. In this paper, inspired by ATDA, we instead design a new \textit{local-descriptor-based unsupervised domain adaptation measure (Local-UDA)} as follows,
\begin{equation}\label{fun2}\small
\begin{split}
  \mathcal{L}_\text{Local-UDA}^\text{fea}(\bm{x}, \bm{x}^\text{adv})= \frac{1}{m}\big\|\bm{\mu}-\bm{\mu}^\text{adv}\big\|_1+
  \frac{1}{m^2}\big\|\bm{\Sigma}-\bm{\Sigma}^\text{adv}\big\|_1\,,
\end{split}
\end{equation}
where $\|\cdot\|_1$ denotes $\ell_1$ norm of a vector or a matrix, and $m$ indicates the number of local descriptors of each image. As seen, like the above proposed KLD and TCD criteria, Local-UDA attempts to align the local-descriptor-based distributions of $\bm{x}$ and $\bm{x}^\text{adv}$ by also taking both the first-order (\textit{i.e.,} mean vectors $\bm{\mu}$ and $\bm{\mu}^\text{adv}$) and second-order (\textit{i.e.,} covariance matrices $\bm{\Sigma}$ and $\bm{\Sigma}^\text{adv}$) information into account.

\subsubsection{Prediction-wise Consistency Criteria}
\label{Section_3_4_3}
In addition to the above feature-wise consistency criteria, we also propose a kind of \textit{prediction-wise consistency criteria} $\mathcal{L}^\text{class}$ to enforce the class predictions of both clean and adversarial examples to be similar, following the existing adversarial defense methods in the literature.

Specifically, we can develop adversarial logit pairing (ALP)~\cite{kannan2018adversarial}, virtual adversarial training (VAT)~\cite{miyato2018virtual} or TRADES~\cite{ZhangICML2019} to achieve this goal. The formulation of ALP can be defined as
\begin{equation}\label{fun6}\small
 \mathcal{L}_\text{ALP}^\text{class}(\bm{x}, \bm{x}^\text{adv})=\|f_\theta\big(g_\varphi(\bm{x}), \mathcal{S}\big)-f_\theta\big(g_\varphi(\bm{x}^\text{adv}), \mathcal{S}\big)\|_2^2\,,
\end{equation}
where $y=f_\theta(g_\varphi(\bm{x}), \mathcal{S})\in\mathbb{R}^C$ and $y^\text{adv}=f_\theta(g_\varphi(\bm{x}^\text{adv}),\mathcal{S})\in\mathbb{R}^C$ denote the posterior probability distributions of class predictions on the clean query image $\bm{x}$ and adversarial query image $\bm{x}^\text{adv}$, respectively. $\|\cdot\|_2$ indicates the $\ell_2$ norm of a vector.

As for VAT~\cite{miyato2018virtual} and TRADES~\cite{ZhangICML2019}, they employ an asymmetric Kullback-Leibler (KL) divergence (for univariate probability distributions) to minimize the difference between the class predictions of the clean example and that of its adversarial counterpart, which can be formulated as,
\begin{equation}\label{fun7}\small
 \mathcal{L}_\text{KL}^\text{class}(\bm{x}, \bm{x}^\text{adv})=\sum_{y,y^\text{adv}\in\mathcal{Y}}f_\theta\big(g_\varphi(\bm{x}), \mathcal{S}\big)\cdot\log{\frac{f_\theta\big(g_\varphi(\bm{x}),\mathcal{S}\big)}{f_\theta\big(g_\varphi(\bm{x}^\text{adv}),\mathcal{S}\big)}}\,,
\end{equation}
where $\mathcal{Y}$ denotes the probability space of $y$ and $y^\text{adv}$.

However, such an asymmetric KL divergence, \textit{i.e.,} $\mathcal{L}_\text{KL}^\text{class}$, will suffer from saturating gradients as claimed in~\cite{kannan2018adversarial}, which may be unstable during optimization. Although ALP~\cite{kannan2018adversarial} has used a symmetric least squares loss (\textit{i.e.,} $\ell_2$ loss) for two images (\textit{i.e.,} $\bm{x}$ and $\bm{x}^\text{adv}$), it is too strict and will easily suffer from the overfitting problem especially in DFSL. As discussed in the second question in Section~\ref{analysis_task}, there is generally a distribution gap between the actual training set (\textit{i.e.,} the auxiliary set $\mathcal{A}$) and the target test set (\textit{i.e.,} the unseen few-shot tasks). If we employ an overly strict regularization criterion at the training stage, we can indeed effectively narrow the distribution gap between the clean and adversarial examples on the training set, but will significantly weaken the model's generalization ability on the diverse target test sets (\textit{e.g.,} some cross-domain scenarios).

\textbf{Symmetric Kullback-Leibler divergence measure (SKL)}. To overcome the above issue, we further propose a \textit{Symmetric Kullback-Leibler divergence measure (SKL)} as 
\begin{equation}\label{fun8}\small
 \mathcal{L}_\text{SKL}^\text{class}(\bm{x}, \bm{x}^\text{adv})=\frac{1}{2}\Big[\mathcal{L}_\text{KL}^\text{class}(\bm{x}, \bm{x}^\text{adv})+\mathcal{L}_\text{KL}^\text{class}(\bm{x}^\text{adv}, \bm{x})\Big]\,.
\end{equation}
The key advantages of $\mathcal{L}_\text{SKL}^\text{class}$ are in two folds: (1) $\mathcal{L}_\text{SKL}^\text{class}$ is still a kind of distribution measure, which only requires the class predictions of $\bm{x}^\text{adv}$ close to the class predictions of $\bm{x}$ and vice versa, rather than forcing their class predictions exactly to be the same. In this sense, $\mathcal{L}_\text{SKL}^\text{class}$ can enjoy a good generalization ability. (2) different from $\mathcal{L}_\text{KL}^\text{class}$, the proposed $\mathcal{L}_\text{SKL}^\text{class}$ is symmetric, which is more stable during optimization. This proposed SKL criterion is regarded as our minor contribution to the topic of DFSL.

Note that $\mathcal{L}_\text{SKL}^\text{class}$ is different from $\mathcal{L}_\text{KLD}^\text{fea}$. The main difference is that $\mathcal{L}_\text{KLD}^\text{fea}$ is built on the intermediate feature representation (\emph{i.e.,} local descriptors obtained from the convolutional feature map) to calculate the difference between the sets of local descriptors from the clean and adversarial examples, while 
$\mathcal{L}_\text{SKL}^\text{class}$ is built on the prediction results to calculate the distance between the class predictions of the clean example and its adversarial counterpart. In other words, $\mathcal{L}_\text{KLD}^\text{fea}$ tries to make the feature representation of the clean example and that of the corresponding adversarial example similar. In contrast, $\mathcal{L}_\text{SKL}^\text{class}$ aims to make the class prediction of the clean example and that of its adversarial counterpart similar.

\subsubsection{Overall Formulation of DFSL}
According to the above analysis, we can define the overall optimization formulation of DFSL in two ways, \textit{i.e.,} feature-wise consistency criteria based overall formulation
\begin{equation}\label{fun9}\small
\begin{split}
 \mathnormal{\Gamma}_\text{overall}^\text{fea} &= \underset{\mathcal{\varphi,\theta}}{\arg\min}\sum_{i=1}^n\sum_{\bm{x}\in \mathcal{Q}_i}\Big(\mathcal{L}^\text{CE}(\bm{x},\mathcal{S}_i,y)\\
 &+ \underset{\bm{x}^\text{adv}=\bm{x}+\bm{\delta}}{\max}\mathcal{L}^\text{CE}(\bm{x}^\text{adv},\mathcal{S}_i,y)+\lambda\cdot\mathcal{L}^\text{fea}(\bm{x}, \bm{x}^\text{adv})\Big)\,,
\end{split}
\end{equation}
or prediction-wise consistency criteria based overall formulation as,
\begin{equation}\label{fun10}\small
\begin{split}
 \mathnormal{\Gamma}_\text{overall}^\text{class} &= \underset{\mathcal{\varphi,\theta}}{\arg\min}\sum_{i=1}^n\sum_{\bm{x}\in \mathcal{Q}_i}\Big(\mathcal{L}^\text{CE}(\bm{x},\mathcal{S}_i,y)\\
 &+ \underset{\bm{x}^\text{adv}=\bm{x}+\bm{\delta}}{\max}\mathcal{L}^\text{CE}(\bm{x}^\text{adv},\mathcal{S}_i,y)+\lambda\cdot\mathcal{L}^\text{class}(\bm{x}, \bm{x}^\text{adv})\Big)\,,
\end{split}
\end{equation}
where $\lambda$ is a balancing parameter.

During the training stage, to learn a specific DFSL model, we can just choose an optimization algorithm, \textit{e.g.,} stochastic gradient descent (SGD) or Adam~\cite{kingma2015adam}, to minimize the above objective $\mathnormal{\Gamma}_\text{overall}$ to optimize the parameters of both the embedding module $g_\varphi$ and classifier module $f_\theta$ from scratch in an end-to-end manner. During the test stage, we directly apply the learned DFSL model, including $g_\varphi$ and $f_\theta$, to the target adversary-based few-shot tasks to classify the images in a query set based on its support set.

\subsection{Generality of the DFSL Framework}
\label{Section_3_5}
We highlight that the proposed DFSL framework is a general framework. This framework consists of two explicitly decoupled modules, \textit{i.e.,} an embedding modules $g_\varphi$ and a classifier module $f_\theta$, and thus most existing few-shot learning (FSL) methods, such as ProtoNet~\cite{snell2017prototypical}, RelationNet~\cite{sung2018learning} and DN4~\cite{li2019DN4}, can be easily tailored into this framework. To be specific, we can just replace the classifier module $f_\theta$ with the corresponding few-shot classifier of different FSL methods, by using the same embedding module $g_\varphi$. Similarly, because of such a decoupled architecture of this framework, either the feature-wise consistency criteria or the prediction-wise consistency criteria (\textit{i.e.,} adversarial defense criteria) can be easily integrated into this DFSL framework. In summary, to construct a specific DFSL model, we can just select a specific FSL method and a specific adversarial defense criterion, and tailor them to this framework.

Typically, to demonstrate the effectiveness of the proposed DFSL framework, we take DN4~\cite{li2019DN4}, one of the state of the arts, as the default FSL method, and compare different adversarial defense methods (criteria). Specifically, five state-of-the-art generic adversarial defense methods, including the standard adversarial training (AT)~\cite{goodfellow2014explaining}, VAT~\cite{miyato2018virtual}, ALP~\cite{kannan2018adversarial}, ATDA~\cite{MMDICLR2019} and TRADES~\cite{ZhangICML2019}, are modified and re-implemented into this unified framework. Based on the methods proposed in Sections~\ref{Section_3_4_2} and \ref{Section_3_4_3}, we can obtain four new DFSL models, \textit{i.e.,} \textit{DFSL-DN4-AT}, \textit{DFSL-DN4-KL}, \textit{DFSL-DN4-ALP} and \textit{DFSL-DN4-Local-UDA}. Note that \textit{DFSL-DN4-Local-UDA} essentially can be seen as our own implementation, inspired by ATDA~\cite{MMDICLR2019}. Moreover, we further construct another three DFSL models, \textit{i.e.,} \textit{DFSL-DN4-KLD (ours)}, \textit{DFSL-DN4-TCD (ours)} and \textit{DFSL-DN4-SKL (ours)}, by using our proposed KLD, TCD and SKL criteria, respectively.

In addition, we have also tailored another five representative few-shot learning methods to this DFSL framework, including ProtoNet~\cite{snell2017prototypical}, RelationNet~\cite{sung2018learning}, CovaMNet~\cite{li2019CovaMNet}, CAN~\cite{CAN_NeurIPS2019} and DeepEMD~\cite{deepemd_CVPR2020}. This part will be discussed in detail in Section~\ref{Section_5_2}.

\section{A New Unified Evaluation Criterion}
 Training a model with adversarial examples can indeed improve the robustness of this model. At the same time, it could jeopardise the performance of this model on the original clean examples. There may exist a trade-off between robustness (against adversarial examples) and accuracy (on clean examples). Several recent works have been trying to show this issue~\cite{tsipras2019robustness,ZhangICML2019}. In the literature, adversarial training based work generally reports two kinds of classification accuracy, \textit{i.e.}, accuracy on clean examples and accuracy on adversarial examples. However, in real cases, it could be awkward to compare two defense methods overall with two accuracies\footnote{If one method achieves a \textit{higher} adversarial accuracy at the expense of a \textit{lower} clean accuracy, we cannot say it is better than another with a \textit{lower} adversarial accuracy but a \textit{higher} clean accuracy.}. To this end, inspired by the case of Recall and Precision in information retrieval, we introduce $\mathcal{F}_{\beta}$ score as a unified criterion to evaluate different defense methods under the same principle. Specifically, $\mathcal{F}_{\beta}$ score is formulated with both clean accuracy $\mathcal{ACC}_\text{clean}$ and adversarial accuracy $\mathcal{ACC}_\text{adv}$ as below,
\begin{equation}\label{fun11}\small
 \mathcal{F}_{\beta}=(1+\beta^2)\cdot\frac{\mathcal{ACC}_\text{clean}\cdot\mathcal{ACC}_\text{adv}}{\beta^2\cdot\mathcal{ACC}_\text{clean}+\mathcal{ACC}_\text{adv}}\,.
\end{equation}
For instance, if we would like to maintain high accuracy on the clean examples, and meanwhile, improve the robustness as higher as possible, we can use $\mathcal{F}_{0.5}$ score. In contrast, if we mainly concern the robustness, $\mathcal{F}_2$ score can be adopted. Similarly, we can use $\mathcal{F}_1$ score to consider these two parts equally. In addition, we can also generate the curve of $\mathcal{F}_{\beta}$ scores by varying $\beta$ as a more comprehensive way to compare different methods.

\section{Experiments}
\label{experiments}
In this section, we perform defensive few-shot image classification on six benchmark datasets to demonstrate the effectiveness of the proposed DFSL framework.

\textbf{Datasets.} Following the literature, six datasets are used as the benchmark datasets, \textit{i.e.}, \emph{mini}ImageNet~\cite{VinyalsBLKW16}, \textit{tiered}ImageNet~\cite{ren2018meta}, CIFAR-100~\cite{krizhevsky2009learning}, Stanford Dogs~\cite{khosla2011novel}, Stanford Cars~\cite{krause20133d} and CUB Birds-200-2011~\cite{wah2011caltech}. \textbf{\emph{mini}ImageNet} consists of $100$ classes, and there are $600$ images in each class with a resolution of $84\!\times\!84$. We follow~\cite{ravi2017optimization} and take $64$, $16$ and $20$ classes for training, validation and test, respectively. \textbf{\textit{tiered}ImageNet} contains $608$ classes with more than $1000$ images per class. We follow the splits in~\cite{ren2018meta} and take $351$, $97$ and $160$ classes for training, validation and test, respectively. \textbf{CIFAR-100} has $100$ classes, containing $600$ images per class. For this dataset, we follow~\cite{sun2019meta} and take $60$, $20$ and $20$ classes for training, validation and test, respectively. \textbf{Stanford Dogs} is a fine-grained dog dataset, which has $120$ classes of dogs and has a total number of $20,580$ images. We follow~\cite{li2019CovaMNet} and take $70$, $20$ and $30$ classes for training, validation and test, respectively. \textbf{Stanford Cars} is a fine-grained car dataset, which contains $196$ classes of cars and has a total number of $16,185$ images. Following~\cite{li2019CovaMNet}, we take $130$, $17$ and $49$ classes for training, validation and test, respectively. \textbf{CUB Birds-200-2011} is a fine-grained bird dataset. It consists of $200$ bird classes containing a total number of $11,788$ images. We also follow~\cite{li2019CovaMNet} and adopt $130$, $20$ and $50$ classes for training, validation and test, respectively. Note that all the images in the above datasets are resized to a resolution of $84\!\times\!84$.

\textbf{Network Architecture.} A commonly used four-layer CNN in generic few-shot learning~\cite{yang2018learning,li2019DN4} is adopted as the embedding module $g_\varphi$. It consists of four convolutional blocks, each of which contains a convolutional layer, a batch normalization layer, and a LeakyReLU layer. As for the classifier module $f_\theta$, it is associated with the selected FSL method. For example, if we choose ProtoNet~\cite{snell2017prototypical} as the FSL method, there will be a fully-connected layer or a global average pooling layer in $f_\theta$. Also, if we choose RelationNet~\cite{sung2018learning} as the FSL method, $f_\theta$ is consists of two convolutional blocks and two fully-connected layers. In contrast, if we adopt the DN4~\cite{li2019DN4} as the base FSL method, the classifier module $f_\theta$ consists of an image-to-class module and a nearest neighbor classifier, which does not have trainable parameters, \textit{i.e.,} $f_\theta$ is non-parametric.

\textbf{Attack Setting.} We apply the popular FGSM~\cite{goodfellow2014explaining} attacker to find adversarial examples to train robust DFSL models. During training, we follow~\cite{kurakin2016adversarial} and randomly choose a perturbation $\epsilon$ for each training few-shot task from a normal distribution in the range of $[0, 0.02]$. To be specific, for each input clean image (\textit{i.e.}, query image), we construct its corresponding one adversarial counterpart based on a specific $\epsilon$. The hype-parameter $\lambda$ in Eq.~(\ref{fun9}) or Eq.~(\ref{fun10}) is selected from $\lambda=\{0.1, 0.5, 1.0, 2.0, 3.0\}$ by cross-validation according to the validation set for each defense method. During test, we evaluate the robustness of the trained DFSL models in defense of three levels of attacks, \textit{i.e.}, $\epsilon=\{0.003, 0.007, 0.01\}$. Basically, the larger $\epsilon$, the stronger the attack.

\textbf{Experimental Setting.} 
For fairness, all experiments are conducted around a $5$-way $5$-shot (or 1-shot) task on the benchmark datasets. At the training stage, we train all the models for $20$ epochs, and in each epoch, we randomly sample and construct $10,000$ adversary-based few-shot tasks. In each conducted few-shot task, there are $5$ support classes with $5$ support images (or $1$ support image) and $15$ query images per class. Adam algorithm~\cite{kingma2015adam} is used to update all the models. The initial learning rate is set as $0.001$ and halved per $10$ epochs. All the models are trained from scratch in an end-to-end manner. At the test stage, $5000$ adversary-based few-shot tasks are constructed from the test set for evaluation. The top-$1$ mean accuracy and the proposed $\mathcal{F}_{\beta}$ score are taken as the evaluation criteria. Without loss of generality, we set $\beta\!=\!1$, which means that the clean accuracy (the weight is $1/2$) is regarded as equally important as the adversarial accuracy (the weight is $1/2$).

\textbf{Experimental Summary.} 
In Section~\ref{Section_5_1}, we will first conduct an experiment on the miniImageNet dataset to verify the effectiveness (\textit{i.e.,} the defense transfer ability) of the proposed \textit{episode-based adversarial training (ET) mechanism}. Next, Section~\ref{Section_5_2} provides details of the generality of the proposed DFSL framework on different few-shot learning methods. In Section~\ref{Section_5_3} and Section~\ref{Section_5_4}, we conduct experiments on three general image datasets and three fine-grained image datasets to compare different DFSL models, respectively. Moreover, the PGD-adversarial training is also introduced to train more robust DFSL models for defending against various attacks in Section~\ref{Section_5_8}.
In Section~\ref{Section_5_5}, we further conduct an experiment in cross-domain scenarios to verify the defense transfer ability across domains of the proposed DFSL framework. In Section~\ref{Section_5_6}, we show the important impact of the randomness. The qualitative comparison, \textit{i.e.,} curves of $\mathcal{F}_{\beta}$ scores, between different DFSL models is given in Section~\ref{Section_5_7}.

According to the analyses and results in Section~\ref{Section_5_6}, we know that randomness indeed matters. That is to say, randomness will make the comparison between different DFSL models unfair. Therefore, for fairness, we will completely fix the randomness for all comparison DFSL models, including implementing all the models with the same single codebase. 

\begin{table*}[!tp]
	\centering
    \extrarowheight=-1pt
	\caption{Comparison ($\%$) of DN4, DN4+AT, DN4+Ep and DN4+ET on \emph{mini}ImageNet under the $5$-way $5$-shot setting. A FGSM attacker is used.}
	 \vspace{-3mm}
	\begin{tabular}{@{}lcc|c|cc|cc|cc@{}}
	\toprule
    \multirow{2}{*}{Method} &\multirow{2}{*}{AT} &\multirow{2}{*}{Ep} &\multirow{2}{*}{$\mathcal{ACC}_\text{clean}$}   &\multicolumn{2}{c|}{$\epsilon\!=\!0.003$}  &\multicolumn{2}{c|}{$\epsilon\!=\!0.007$}  &\multicolumn{2}{c}{$\epsilon\!=\!0.01$} \\
                           & & & &$\mathcal{ACC}_\text{adv}$ &$\mathcal{F}_1$  &$\mathcal{ACC}_\text{adv}$ &$\mathcal{F}_1$ &$\mathcal{ACC}_\text{adv}$ &$\mathcal{F}_1$\\
    \midrule
    DN4       & without & without  &$57.74$      &$33.99$      &$42.79$       &$24.07$       &$33.97$       &$21.54$      &$31.37$ \\
    DN4+AT    & with    & without  &$52.12$      &$44.63$      &$48.08$       &$40.99$       &$45.88$       &$38.19$      &$44.08$ \\
    DN4+Ep    & without & with     &$70.39$      &$16.18$      &$26.31$       &$9.81$        &$17.22$       &$8.65$       &$15.40$ \\
    DN4+ET & with       & with     &$\bm{71.54}$ &$\bm{63.31}$ &$\bm{67.17}$  &$\bm{58.92}$  &$\bm{64.61}$  &$\bm{54.97}$ &$\bm{62.16}$ \\
    \bottomrule
	\end{tabular}
	\label{episodic_result}
	\vspace{-2mm}
\end{table*}
\begin{table*}[!tp]
	\centering
    \extrarowheight=-1pt
	\caption{Generality of the proposed DFSL framework with adversarial training (AT)~\cite{goodfellow2014explaining} to different FSL methods, including ProtoNet~\cite{SnellSZ17Prototypical}, RelationNet~\cite{sung2018learning}, CovaMNet~\cite{li2019CovaMNet}, CAN~\cite{CAN_NeurIPS2019}, DeepEMD~\cite{deepemd_CVPR2020} and DN4~\cite{li2019DN4}. A FGSM attacker is used on \emph{mini}ImageNet under the $5$-way $5$-shot setting.}
	 \vspace{-3mm}
	\begin{tabular}{@{}lcc|c|cc|cc|cc@{}}
	\toprule
    \multirow{2}{*}{Method} &\multirow{2}{*}{ET}  &\multirow{2}{*}{Feature} &\multirow{2}{*}{$\mathcal{ACC}_\text{clean}$}   &\multicolumn{2}{c|}{$\epsilon\!=\!0.003$}  &\multicolumn{2}{c|}{$\epsilon\!=\!0.007$}  &\multicolumn{2}{c}{$\epsilon\!=\!0.01$} \\
                            & & & &$\mathcal{ACC}_\text{adv}$ &$\mathcal{F}_1$  &$\mathcal{ACC}_\text{adv}$ &$\mathcal{F}_1$ &$\mathcal{ACC}_\text{adv}$ &$\mathcal{F}_1$\\
    \midrule
    ProtoNet+AT             &without  &Global  &$54.61$      &$48.81$      &$51.54$       &$41.51$       &$47.16$       &$36.58$      &$43.81$ \\
    DeepEMD+AT &without  &Local   &$55.09$      &$49.13$      &$51.93$       &$42.06$       &$47.70$       &$37.61$      &$44.70$ \\
    DN4+AT                  &without  &Local   &$52.12$      &$44.63$      &$48.08$       &$40.99$       &$45.88$       &$38.19$      &$44.08$ \\
    \midrule
    DFSL-ProtoNet-AT        &with     &Global  &$63.70$      &$55.68$      &$59.42$       &$45.91$       &$53.36$       &$39.05$      &$48.41$ \\
    DFSL-RelationNet-AT     &with     &Global  &$62.45$      &$54.94$      &$58.45$       &$46.04$       &$53.00$       &$39.86$      &$48.66$ \\
    DFSL-DeepEMD-AT        &with     &Local   &$60.20$      &$52.65$      &$56.17$       &$43.97$       &$50.82$       &$38.35$      &$46.85$ \\
    DFSL-CAN-AT            &with     &Local   &$64.64$      &$56.04$      &$60.03$       &$45.96$       &$53.72$       &$38.83$      &$48.51$ \\
    DFSL-CovaMNet-AT        &with     &Local   &$64.65$      &$57.57$      &$60.90$       &$49.19$       &$55.87$       &$43.10$      &$51.72$ \\
    DFSL-DN4-AT             &with     &Local   &$\bm{71.54}$ &$\bm{63.31}$ &$\bm{67.17}$  &$\bm{58.92}$  &$\bm{64.61}$  &$\bm{54.97}$ &$\bm{62.16}$ \\
    \bottomrule
	\end{tabular}
	\label{fsl_results}
	\vspace{-2mm}
\end{table*}

\subsection{Verifying the Defense Transfer Ability of ET}
\label{Section_5_1}
One of our concerns is how to transfer the adversarial defense knowledge from a sample distribution to another different one. We propose an \textit{episode-based adversarial training (ET) mechanism} to achieve this. To verify the effectiveness of ET, we conduct a comparison experiment on \textit{mini}ImageNet, built on one state-of-the-art FSL method DN4~\cite{li2019DN4}. Specifically, four variants of DN4 are constructed by considering whether to perform the standard adversarial training (AT)~\cite{goodfellow2014explaining} and whether to perform the standard episodic training mechanism (Ep for short)~\cite{VinyalsBLKW16}. They are DN4, DN4+AT, DN4+Ep, and DN4+ET (\textit{i.e.,} DN4+Ep+AT).

The first variant, \textit{i.e.}, DN4, is trained without any of AT and Ep. To achieve this, we first train a $64$-classes classification network on the auxiliary set $\mathcal{A}$ (64 classes) of \textit{mini}ImageNet, and then, based on this pre-trained network, we directly perform the non-parametric classifier module of the original DN4 in~\cite{li2019DN4} on the test set (20 classes). DN4+AT is trained in a similar way, but with additional adversarial training (AT) at the training stage of the $64$-classes classification network. DN4+Ep is trained on the auxiliary set $\mathcal{A}$ by using Ep like the original work in~\cite{li2019DN4} but without using any adversarial defense technique. As for DN4+ET, we train it on $\mathcal{A}$ with the proposed episode-based adversarial training (ET), and evaluate it on the test set.

All the results are reported in Table~\ref{episodic_result}. As seen, without AT, DN4 is much vulnerable to adversarial attacks, dropping its accuracy from $57.74\%$ (clean accuracy) to $21.54\%$ (adversarial accuracy) when $\epsilon=0.01$. In contrast, DN4+AT, which is trained with AT, can indeed defend the adversarial attacks as expected (from $21.54\%$ to $38.19\%$). However, both clean and adversarial accuracies of DN4+AT are still far from the normal accuracy on \textit{mini}ImageNet ($70.39\%$).

Fortunately, the episodic training makes DN4+ET perform much better than DN4+AT (which is trained without Ep) on both clean and adversarial examples. For instance, when $\epsilon\!=\!0.01$, DN4+ET obtains $19.42\%$ and $16.78\%$ improvements over DN4+AT on the clean accuracy and adversarial accuracy, respectively. More importantly, the clean accuracy ($71.54\%$) is even better than the normal clean accuracy ($70.39\%$). Therefore, it suggests that the episode-based adversarial training (ET) can not only transfer the adversarial defense knowledge but also maintain the clean classification knowledge. This verifies that the traditional sample-level distribution consistency assumption may not guarantee the model's generalization on both clean and adversarial examples, but the task-level distribution consistency assumption during episodic training can properly make it, as stated in the first question in Section~\ref{analysis_task}.

\subsection{Generality of DFSL to Different FSL Methods}
\label{Section_5_2}
To verify the generality of the proposed DFSL framework, we first extend this framework to six representative FSL methods, \textit{i.e.,} ProtoNet~\cite{SnellSZ17Prototypical}, RelationNet~\cite{sung2018learning}, CovaMNet~\cite{li2019CovaMNet}, CAN~\cite{CAN_NeurIPS2019}, DeepEMD~\cite{deepemd_CVPR2020} and DN4~\cite{li2019DN4}, and then compare these models with each other. For fairness, we just select the standard adversarial training (AT)~\cite{goodfellow2014explaining} as the adversarial defense method for these models. 
In addition, note that in the original papers CAN~\cite{CAN_NeurIPS2019} leverages a global classification as an additional training task by using the global true labels of the auxiliary set, and DeepEMD~\cite{deepemd_CVPR2020} employs the pre-training on the auxiliary set as a pre-processing operation. This is somewhat not fair for other methods in comparison and is also not consistent with our setting in DFSL. Therefore, we re-implement CAN and DeepEMD with their core components by removing the global classification part or the pre-training part. In this sense, we can obtain six corresponding DFSL models, \textit{i.e.,} \textit{DFSL-ProtoNet-AT}, \textit{DFSL-RelationNet-AT}, \textit{DFSL-CovaMNet-AT}, \textit{DFSL-CAN-AT}, \textit{DFSL-DeepEMD-AT} and \textit{DFSL-DN4-AT}.

Moreover, to show the effectiveness of the above models, we take ProtoNet+AT, DeepEMD+AT and DN4+AT as the baselines. Specifically, ProtoNet+AT and DeepEMD+AT are trained in the same way as DN4+AT (see Section~\ref{Section_5_1} for more details). Note that, we do not implement RelationNet+AT, CovaMNet+AT and CAN+AT, because their classifier modules have trainable parameters and thus cannot be directly integrated with an adversarially pre-trained embedding network like ProtoNet+AT, DeepEMD+AT and DN4+AT at the test stage without fine-tuning. The comparison results of the above models on \textit{mini}ImageNet are reported in Table~\ref{fsl_results}. As seen, all the six DFSL-based methods can significantly improve both the clean accuracy and adversarial accuracy over ProtoNet+AT, DeepEMD+AT and DN4+AT. This verifies the generality and effectiveness of the proposed DFSL framework. In addition, it also verifies that the existing adversarial method (\emph{e.g.,} AT) indeed should be combined with the episodic training mechanism to make FSL methods more robust.

As aforementioned, we prefer to use DN4~\cite{li2019DN4} as the default FSL method, because DN4 is one of the state-of-the-art FSL methods. From Table~\ref{fsl_results}, we can see that DFSL-DN4-AT indeed performs significantly better than the other five models on both the clean and adversarial accuracies. For example, on the clean accuracy, DFSL-DN4-AT gains $7.84\%$, $9.09\%$, $11.34\%$, $6.90\%$ and $6.89\%$ improvements over DFSL-ProtoNet-AT, DFSL-RelationNet-AT, DFSL-DeepEMD-AT, DFSL-CAN-AT, and DFSL-CovaMNet-AT, respectively. Also, on the adversarial accuracy ($\epsilon\!=\!0.01$), DFSL-DN4-AT obtains $15.92\%$, $15.11\%$, $16.62\%$, $16.07\%$, and $11.87\%$ improvements over these methods, respectively. This is because DN4~\cite{li2019DN4} employs an image-to-class module to perform the final classification, enjoying the exchangeability of local patterns inside each class, which can significantly improve the robustness of a DFSL model against the adversarial perturbations. Therefore, in the remaining experiments, we will employ DN4 as the default FSL method of our DFSL framework.

\begin{table*}[!tp] 
	\centering
    \extrarowheight=-1pt
	\caption{Comparison ($\%$) of different DFSL models on \textit{mini}ImageNet under the $5$-way $5$-shot setting. Both training and test are based on a FGSM attacker. For each evaluation criterion, the best and the second best results are highlighted in bold.}
	\vspace{-3mm}
	\begin{tabular}{@{}lc|c|cc|cc|cc@{}}
	\toprule
    \multirow{2}{*}{Method} &\multirow{2}{*}{ET} &\multirow{2}{*}{$\mathcal{ACC}_\text{clean}$}   &\multicolumn{2}{c|}{$\epsilon\!=\!0.003$}  &\multicolumn{2}{c|}{$\epsilon\!=\!0.007$}  &\multicolumn{2}{c}{$\epsilon\!=\!0.01$} \\
                           & &  &$\mathcal{ACC}_\text{adv}$ &$\mathcal{F}_1$  &$\mathcal{ACC}_\text{adv}$ &$\mathcal{F}_1$ &$\mathcal{ACC}_\text{adv}$ &$\mathcal{F}_1$\\
    \midrule
    DN4+Ep                         &without  &$70.39$        &$16.18$       &$26.31$      &$9.81$       &$17.22$        &$8.65$        &$15.40$ \\
    DFSL-DN4-AT                    &with     &$71.54$        &$63.31$       &$67.17$      &$58.92$      &$64.61$        &$54.97$       &$62.16$ \\
    DFSL-DN4-KL                    &with     &$71.61$        &$64.96$       &$68.12$      &$62.29$      &$66.62$        &$59.55$       &$65.02$ \\
    DFSL-DN4-ALP                   &with     &$72.14$        &$64.56$       &$68.13$      &$62.56$      &$67.00$        &$\bm{60.87}$  &$\bm{66.02}$ \\
    DFSL-DN4-Local-UDA             &with     &$\bm{72.30}$   &$64.32$       &$68.07$      &$61.46$      &$66.44$        &$59.07$       &$65.01$ \\
    \textbf{DFSL-DN4-KLD (ours)}   &with     &$72.24$        &$63.38$       &$67.52$      &$61.06$      &$66.18$        &$59.24$       &$65.09$ \\
    \textbf{DFSL-DN4-TCD (ours)}   &with     &$\bm{72.33}$   &$\bm{64.98}$  &$\bm{68.45}$ &$\bm{62.76}$ &$\bm{67.20}$   &$60.65$       &$65.97$ \\
    \textbf{DFSL-DN4-SKL (ours)}   &with     &$71.68$        &$\bm{66.44}$  &$\bm{68.96}$ &$\bm{64.55}$ &$\bm{67.92}$   &$\bm{62.68}$  &$\bm{66.98}$ \\
    \bottomrule
	\end{tabular}
	\label{comparison_FGSM_result_mini}
	\vspace{-3mm}
\end{table*}

\begin{table*}[!tp]
	\centering
    \extrarowheight=-1pt
	\caption{Comparison ($\%$) of different DFSL models on \textit{tiered}ImageNet under the $5$-way $5$-shot setting. Both training and test are based on a FGSM attacker. For each evaluation criterion, the best and the second best results are highlighted in bold.}
	\vspace{-3mm}
	\begin{tabular}{@{}lc|c|cc|cc|cc@{}}
	\toprule
    \multirow{2}{*}{Method} &\multirow{2}{*}{ET} &\multirow{2}{*}{$\mathcal{ACC}_\text{clean}$}   &\multicolumn{2}{c|}{$\epsilon\!=\!0.003$}  &\multicolumn{2}{c|}{$\epsilon\!=\!0.007$}  &\multicolumn{2}{c}{$\epsilon\!=\!0.01$} \\
                           & &  &$\mathcal{ACC}_\text{adv}$ &$\mathcal{F}_1$  &$\mathcal{ACC}_\text{adv}$ &$\mathcal{F}_1$ &$\mathcal{ACC}_\text{adv}$ &$\mathcal{F}_1$\\
    \midrule
    DN4+Ep                        &without &$\bm{74.11}$ &$23.99$       &$36.24$        &$12.58$        &$21.50$       &$10.34$      &$18.14$ \\
    DFSL-DN4-AT                   &with    &$\bm{73.04}$ &$63.06$       &$67.68$        &$58.07$        &$64.70$       &$53.73$      &$61.91$ \\
    DFSL-DN4-KL                   &with    &$72.31$      &$64.60$       &$68.23$        &$61.96$        &$66.76$       &$59.26$      &$65.13$ \\
    DFSL-DN4-ALP                  &with    &$72.61$      &$65.37$       &$68.80$        &$62.84$        &$67.37$       &$60.55$      &$66.03$ \\
    DFSL-DN4-Local-UDA            &with    &$72.48$      &$64.12$       &$68.04$        &$60.65$        &$66.03$       &$57.04$      &$63.83$ \\
    \textbf{DFSL-DN4-KLD (ours)}  &with    &$72.71$      &$\bm{66.58}$  &$\bm{69.51}$   &$\bm{64.27}$   &$\bm{68.23}$  &$\bm{61.89}$  &$\bm{66.86}$ \\
    \textbf{DFSL-DN4-TCD (ours)}  &with    &$72.66$      &$\bm{67.39}$  &$\bm{69.92}$   &$\bm{65.55}$   &$\bm{68.92}$  &$\bm{63.67}$  &$\bm{67.86}$ \\
    \textbf{DFSL-DN4-SKL (ours)}  &with    &$72.43$      &$65.57$       &$68.82$        &$63.73$        &$67.80$       &$61.64$       &$66.60$ \\
    \bottomrule
	\end{tabular}
	\label{comparison_FGSM_result_tier}
	\vspace{-3mm}
\end{table*}

\begin{table*}[!tp]
	\centering
    \extrarowheight=-1pt
	\caption{Comparison ($\%$) of different DFSL models on CIFAR-100 under the $5$-way $5$-shot setting. Both training and test are based on a FGSM attacker. For each evaluation criterion, the best and the second best results are highlighted in bold.}
	\vspace{-3mm}
	\begin{tabular}{@{}lc|c|cc|cc|cc@{}}
	\toprule
    \multirow{2}{*}{Method} &\multirow{2}{*}{ET} &\multirow{2}{*}{$\mathcal{ACC}_\text{clean}$}   &\multicolumn{2}{c|}{$\epsilon\!=\!0.003$}  &\multicolumn{2}{c|}{$\epsilon\!=\!0.007$}  &\multicolumn{2}{c}{$\epsilon\!=\!0.01$} \\
                           & &  &$\mathcal{ACC}_\text{adv}$ &$\mathcal{F}_1$  &$\mathcal{ACC}_\text{adv}$ &$\mathcal{F}_1$ &$\mathcal{ACC}_\text{adv}$ &$\mathcal{F}_1$\\
    \midrule
    DN4+Ep                        &without &$\bm{59.93}$  &$9.18$    &$15.92$   &$6.87$    &$12.32$    &$6.70$     &$12.05$ \\
    DFSL-DN4-AT                   &with    &$55.03$       &$44.88$   &$49.43$   &$43.30$   &$48.46$    &$41.67$    &$47.42$ \\
    DFSL-DN4-KL                   &with    &$54.84$       &$46.15$   &$50.12$   &$44.17$   &$48.93$    &$42.37$    &$47.80$ \\
    DFSL-DN4-ALP                  &with    &$55.59$       &$45.25$   &$49.88$   &$42.07$   &$47.89$    &$40.59$    &$46.92$ \\
    DFSL-DN4-Local-UDA            &with    &$55.26$       &$45.94$   &$50.17$   &$43.89$   &$48.92$    &$42.22$    &$47.86$ \\
    \textbf{DFSL-DN4-KLD (ours)}  &with    &$\bm{56.12}$ &$\bm{46.38}$  &$\bm{50.78}$   &$\bm{44.86}$  &$\bm{49.86}$  &$\bm{43.39}$ &$\bm{48.94}$ \\
    \textbf{DFSL-DN4-TCD (ours)}  &with    &$54.77$      &$45.85$       &$49.91$        &$43.89$       &$48.73$       &$42.08$      &$47.59$ \\
    \textbf{DFSL-DN4-SKL (ours)}  &with    &$55.51$      &$\bm{48.88}$  &$\bm{51.98}$   &$\bm{46.37}$  &$\bm{50.53}$  &$\bm{44.79}$ &$\bm{49.57}$ \\
    \bottomrule
	\end{tabular}
	\label{comparison_FGSM_result_cifar}
	\vspace{-3mm}
\end{table*}

\subsection{Performing DFSL on General Image Datasets}
\label{Section_5_3}
In this section, we perform DFSL on three general image datasets, \textit{i.e.,} \emph{mini}ImageNet, \emph{tiered}ImageNet and CIFAR-100, to further verify the effectiveness of this framework. Note that we don't perform data augmentation on \emph{mini}ImageNet and \emph{tiered}ImageNet. We only perform simple data augmentation on CIFAR-100 because it has a much lower resolution. In addition, because there are no existing methods developed in the literature for the proposed problem, \textit{defensive few-shot learning (DFSL)}, we modify and re-implement the general state-of-the-art adversarial defense methods into the DFSL framework as the benchmarks. This also can be seen as our one minor contribution to the topic of DFSL.

Specifically, as mentioned in Section~\ref{Section_3_5}, seven DFSL-based models will be compared with each other, including \textit{DFSL-DN4-AT}, \textit{DFSL-DN4-KL}, \textit{DFSL-DN4-ALP} and \textit{DFSL-DN4-Local-UDA}, \textit{DFSL-DN4-KLD (ours)}, \textit{DFSL-DN4-TCD (ours)} and \textit{DFSL-DN4-SKL (ours)}. \textit{DN4+Ep} is taken as the baseline, which is just an episodically trained DN4~\cite{li2019DN4} without using any adversarial defense technique. The results on \emph{mini}ImageNet, \emph{tiered}ImageNet and CIFAR-100 are reported in Table~\ref{comparison_FGSM_result_mini}, Table~\ref{comparison_FGSM_result_tier} and Table~\ref{comparison_FGSM_result_cifar}, respectively.

From the results, we can see that all the DFSL models can significantly improve the adversarial accuracy over DN4+Ep on all three levels of adversarial attacks. This further verifies that the proposed \textit{episode-based adversarial training (ET)} can effectively defend against the adversarial attacks. As feature-wise consistency based models, both DFSL-DN4-KLD (ours) and DFSL-DN4-TCD (ours) can dramatically improve the clean accuracy. For example, on \emph{mini}ImageNet in Table~\ref{comparison_FGSM_result_mini}, DFSL-DN4-TCD (ours) can achieve the best clean accuracy ($72.33\%$), which is even much higher than DN4+Ep ($70.39\%$) which is specially trained on the clean examples. Analogously, DFSL-DN4-KLD (ours) obtain the second best clean accuracy ($56.12\%$) on CIFAR-100 in Table~\ref{comparison_FGSM_result_cifar}, which gains $1.09\%$, $1.28\%$, $0.53\%$ and $0.86\%$ improvement over DFSL-DN4-AT, DFSL-DN4-KL, DFSL-DN4-ALP and DFSL-DN4-Local-UDA, respectively. This is because KLD and TCD are distribution measure based criteria, which can effectively leverage the local-descriptor-based distributions of both clean and adversarial images. Notably, as explained in Section~\ref{Section_3_4_2}, TCD encodes the task information with a covariance matrix of the entire support set of the current task, which is able to adapt to diverse adversary-based few-shot tasks.

\begin{table*}[!tp]
	\centering
    \extrarowheight=-1pt
	\caption{\textcolor{black}{Comparison ($\%$) of different DFSL models on \textit{mini}ImageNet under the $5$-way $1$-shot setting. Both training and test are based on a FGSM attacker. For each evaluation criterion, the best and the second best results are highlighted in bold.}}
	\vspace{-3mm}
	{\color{black}\begin{tabular}{@{}lc|c|cc|cc|cc@{}}
	\toprule
    \multirow{2}{*}{Method} &\multirow{2}{*}{ET} &\multirow{2}{*}{$\mathcal{ACC}_\text{clean}$}   &\multicolumn{2}{c|}{$\epsilon\!=\!0.003$}  &\multicolumn{2}{c|}{$\epsilon\!=\!0.007$}  &\multicolumn{2}{c}{$\epsilon\!=\!0.01$} \\
                           & &  &$\mathcal{ACC}_\text{adv}$ &$\mathcal{F}_1$  &$\mathcal{ACC}_\text{adv}$ &$\mathcal{F}_1$ &$\mathcal{ACC}_\text{adv}$ &$\mathcal{F}_1$\\
    \midrule
    DN4+Ep                         &without  &$50.57$        &$22.73$       &$31.36$      &$17.01$      &$25.45$        &$14.45$       &$22.47$ \\
    DFSL-DN4-AT                    &with     &$50.35$        &$44.43$       &$47.20$      &$42.63$      &$46.16$        &$41.01$       &$45.20$ \\
    DFSL-DN4-KL                    &with     &$50.65$        &$45.12$       &$47.72$      &$\bm{43.66}$ &$46.89$        &$\bm{42.32}$  &$46.11$ \\
    DFSL-DN4-ALP                   &with     &$50.32$        &$43.96$       &$46.92$      &$41.29$      &$45.36$        &$39.30$       &$44.13$ \\
    DFSL-DN4-Local-UDA             &with     &$51.08$        &$45.06$       &$47.88$      &$43.58$      &$47.03$        &$42.14$       &$46.18$ \\
    \textbf{DFSL-DN4-KLD (ours)}   &with     &$\bm{51.96}$   &$\bm{46.10}$  &$\bm{48.85}$ &$43.57$      &$\bm{47.39}$   &$42.12$       &$\bm{46.52}$ \\
    \textbf{DFSL-DN4-TCD (ours)}   &with     &$50.82$        &$44.56$       &$47.48$      &$42.99$      &$46.57$        &$41.73$       &$45.82$ \\
    \textbf{DFSL-DN4-SKL (ours)}   &with     &$\bm{52.00}$   &$\bm{46.67}$  &$\bm{49.19}$ &$\bm{44.16}$ &$\bm{47.76}$   &$\bm{42.46}$  &$\bm{46.74}$ \\
    \bottomrule
	\end{tabular}}
	\label{comparison_FGSM_result_mini2}
	\vspace{-3mm}
\end{table*}

\begin{table*}[!tp]
	\centering
    \extrarowheight=-1pt
	\caption{\textcolor{black}{Comparison ($\%$) of different DFSL models on \textit{tiered}ImageNet under the $5$-way $1$-shot setting. Both training and test are based on a FGSM attacker. For each evaluation criterion, the best and the second best results are highlighted in bold.}}
	\vspace{-3mm}
	{\color{black}\begin{tabular}{@{}lc|c|cc|cc|cc@{}}
	\toprule
    \multirow{2}{*}{Method} &\multirow{2}{*}{ET} &\multirow{2}{*}{$\mathcal{ACC}_\text{clean}$}   &\multicolumn{2}{c|}{$\epsilon\!=\!0.003$}  &\multicolumn{2}{c|}{$\epsilon\!=\!0.007$}  &\multicolumn{2}{c}{$\epsilon\!=\!0.01$} \\
                           & &  &$\mathcal{ACC}_\text{adv}$ &$\mathcal{F}_1$  &$\mathcal{ACC}_\text{adv}$ &$\mathcal{F}_1$ &$\mathcal{ACC}_\text{adv}$ &$\mathcal{F}_1$\\
    \midrule
    DN4+Ep                        &without &$\bm{53.76}$ &$14.43$       &$22.75$        &$4.68$         &$8.61$        &$3.01$        &$5.70$ \\
    DFSL-DN4-AT                   &with    &$49.27$      &$43.87$       &$46.41$        &$\bm{42.27}$   &$\bm{45.50}$  &$\bm{40.51}$  &$44.46$ \\
    DFSL-DN4-KL                   &with    &$49.04$      &$42.93$       &$45.78$        &$41.73$        &$45.09$       &$40.30$       &$44.24$ \\
    DFSL-DN4-ALP                  &with    &$49.85$      &$\bm{44.10}$  &$\bm{46.79}$   &$41.33$        &$45.19$       &$39.28$       &$43.93$ \\
    DFSL-DN4-Local-UDA            &with    &$49.00$      &$43.55$       &$46.11$        &$41.67$        &$45.03$       &$39.44$       &$43.70$ \\
    \textbf{DFSL-DN4-KLD (ours)}  &with    &$49.60$      &$43.39$       &$46.28$        &$41.91$        &$45.43$       &$40.44$       &$\bm{44.55}$ \\
    \textbf{DFSL-DN4-TCD (ours)}  &with    &$49.23$      &$42.71$       &$45.73$        &$41.43$        &$44.99$       &$40.06$       &$44.17$ \\
    \textbf{DFSL-DN4-SKL (ours)}  &with    &$\bm{50.06}$ &$\bm{44.88}$  &$\bm{47.32}$   &$\bm{43.24}$   &$\bm{46.40}$  &$\bm{41.00}$  &$\bm{45.07}$ \\
    \bottomrule
	\end{tabular}}
	\label{comparison_FGSM_result_tier2}
	\vspace{-3mm}
\end{table*}

\begin{table*}[!tp]
	\centering
    \extrarowheight=-1pt
	\caption{\textcolor{black}{Comparison ($\%$) of different DFSL models on CIFAR-100 under the $5$-way $1$-shot setting. Both training and test are based on a FGSM attacker. For each evaluation criterion, the best and the second best results are highlighted in bold.}}
	\vspace{-3mm}
	{\color{black}\begin{tabular}{@{}lc|c|cc|cc|cc@{}}
	\toprule
    \multirow{2}{*}{Method} &\multirow{2}{*}{ET} &\multirow{2}{*}{$\mathcal{ACC}_\text{clean}$}   &\multicolumn{2}{c|}{$\epsilon\!=\!0.003$}  &\multicolumn{2}{c|}{$\epsilon\!=\!0.007$}  &\multicolumn{2}{c}{$\epsilon\!=\!0.01$} \\
                           & &  &$\mathcal{ACC}_\text{adv}$ &$\mathcal{F}_1$  &$\mathcal{ACC}_\text{adv}$ &$\mathcal{F}_1$ &$\mathcal{ACC}_\text{adv}$ &$\mathcal{F}_1$\\
    \midrule
    DN4+Ep                        &without &$37.55$       &$4.37$        &$7.82$        &$1.27$        &$2.45$       &$0.93$        &$1.81$ \\
    DFSL-DN4-AT                   &with    &$38.29$       &$31.51$       &$34.57$       &$25.17$       &$30.37$      &$21.26$       &$27.33$ \\
    DFSL-DN4-KL                   &with    &$38.71$       &$32.66$       &$35.42$       &$31.37$       &$34.65$      &$30.18$       &$33.91$ \\
    DFSL-DN4-ALP                  &with    &$38.46$       &$31.92$       &$34.88$       &$25.64$       &$30.76$      &$21.68$       &$27.72$ \\
    DFSL-DN4-Local-UDA            &with    &$38.20$       &$31.54$       &$34.55$       &$25.16$       &$30.38$      &$21.26$       &$27.31$ \\
    \textbf{DFSL-DN4-KLD (ours)}  &with    &$\bm{38.77}$  &$\bm{33.28}$  &$\bm{35.81}$  &$\bm{32.06}$  &$\bm{35.09}$ &$\bm{30.82}$  &$\bm{34.34}$ \\
    \textbf{DFSL-DN4-TCD (ours)}  &with    &$38.13$       &$31.83$       &$34.68$       &$25.57$       &$30.61$      &$21.77$       &$27.71$ \\
    \textbf{DFSL-DN4-SKL (ours)}  &with    &$\bm{39.41}$  &$\bm{33.83}$  &$\bm{36.40}$  &$\bm{32.89}$  &$\bm{35.85}$ &$\bm{31.98}$  &$\bm{35.30}$ \\
    \bottomrule
	\end{tabular}}
	\label{comparison_FGSM_result_cifar2}
	\vspace{-3mm}
\end{table*}

We can also observe that the prediction-wise consistency based methods, such as DFSL-DN4-KL, DFSL-DN4-ALP and DFSL-DN4-SKL (ours), tend to significantly improve the adversarial accuracy, compared to DFSL-DN4-AT. More importantly, the proposed DFSL-DN4-SKL (ours) can further improve the adversarial accuracy over DFSL-DN4-KL, DFSL-DN4-ALP. For example, on the \emph{mini}ImageNet dataset in Table~\ref{comparison_FGSM_result_mini}, when $\epsilon=0.01$, DFSL-DN4-SKL (ours) further gains $3.13\%$ and $1.81\%$ improvements over DFSL-DN4-KL, DFSL-DN4-ALP, respectively. Also, on the CIFAR-100 dataset in Table~\ref{comparison_FGSM_result_cifar}, when $\epsilon=0.01$, such further improvements are $2.42\%$ and $4.2\%$, respectively. As explained in Section~\ref{Section_3_4_3}, the reason is that the proposed SKL is symmetric which is more stable than the asymmetric KL~\cite{miyato2018virtual,ZhangICML2019}. On the other hand, SKL is somewhat slack, and thus it enjoys a good generalization ability.

In addition to the clean accuracy and adversarial accuracy, we also calculate $\mathcal{F}_1$ scores for each model by considering both the clean accuracy and adversarial accuracy together. To explain the necessity of this unified evaluation criterion, we can pay attention to a specific example. On the CIFAR-100 dataset, we notice that DFSL-DN4-ALP has higher clean accuracy ($55.59\%$) than that ($54.84\%$) of DFSL-DN4-KL, but obtains a lower adversarial accuracy ($45.25\%$ when $\epsilon=0.003$) than that ($46.15\%$ when $\epsilon=0.003$) of DFSL-DN4-KL. In this case, it is difficult to conclude which one of these two models is better. In contrast, by calculating a unified $\mathcal{F}_1$ score for each of these models, we can concretely draw a conclusion that DFSL-DN4-KL ($\mathcal{F}_1=50.12\%$) performs better than DFSL-DN4-ALP ($\mathcal{F}_1=49.88\%$). Notably, the proposed methods (\textit{i.e.,} KLD, TCD and SKL) can obtain the highest $\mathcal{F}_1$ scores on all three levels of adversarial attacks on all three datasets.

Furthermore, as seen in Tables~\ref{comparison_FGSM_result_mini2},~\ref{comparison_FGSM_result_tier2} and~\ref{comparison_FGSM_result_cifar2}, we also perform all the DFSL models on the three general image datasets under the $5$-way $1$-shot setting, respectively. From Table~\ref{comparison_FGSM_result_mini2}, we can see that the seven DFSL models consistently improve the adversarial accuracy over DN4+Ep, and the proposed variants with KLD, TCD and SKL can further boost the clean accuracy or adversarial accuracy over other competitors. Notably, the proposed DFSL-DN4-SKL (ours) consistently performs best on both the clean accuracy and adversarial accuracy at all the three levels of adversarial attacks. Moreover, the proposed DFSL-DN4-KLD (ours) also shows competitive performance in most of cases. The proposed DFSL-DN4-TCD (ours) is not so competitive as SKL and KLD. This is because it is truly challenging to effectively estimate the covariance matrix of the support set under the $1$-shot setting (\emph{i.e.}, only one sample is available for estimation) for TCD. In addition, in Tables~\ref{comparison_FGSM_result_tier2} and~\ref{comparison_FGSM_result_cifar2}, we can obtain observations similar to Table~\ref{comparison_FGSM_result_mini2}. For example, the proposed DFSL-DN4-SKL (ours) is consistently superior to other competitors.

\begin{table*}[!tp]
	\centering
    \extrarowheight=-1pt
	\caption{Comparison ($\%$) of different DFSL models on Stanford Dogs under the $5$-way $5$-shot setting. Both training and test are based on a FGSM attacker. For each evaluation criterion, the best and the second best results are highlighted in bold.}
	\vspace{-3mm}
	\begin{tabular}{@{}lc|c|cc|cc|cc@{}}
	\toprule
    \multirow{2}{*}{Method} &\multirow{2}{*}{ET} &\multirow{2}{*}{$\mathcal{ACC}_\text{clean}$}   &\multicolumn{2}{c|}{$\epsilon\!=\!0.003$}  &\multicolumn{2}{c|}{$\epsilon\!=\!0.007$}  &\multicolumn{2}{c}{$\epsilon\!=\!0.01$} \\
                           & &  &$\mathcal{ACC}_\text{adv}$ &$\mathcal{F}_1$  &$\mathcal{ACC}_\text{adv}$ &$\mathcal{F}_1$ &$\mathcal{ACC}_\text{adv}$ &$\mathcal{F}_1$\\
    \midrule
    DN4+Ep                          &without &$\bm{75.95}$    &$15.41$        &$25.62$      &$6.89$         &$12.63$    &$5.69$        &$10.58$ \\
    DFSL-DN4-AT                     &with    &$74.63$         &$59.98$        &$66.50$      &$53.62$        &$62.40$    &$49.86$       &$59.78$ \\
    DFSL-DN4-KL                     &with    &$74.12$         &$60.09$        &$66.37$      &$55.11$        &$63.21$    &$51.81$       &$60.98$ \\
    DFSL-DN4-ALP                    &with    &$67.22$         &$55.94$        &$61.06$      &$42.92$        &$52.41$    &$35.63$       &$46.57$ \\
    DFSL-DN4-Local-UDA              &with    &$75.25$         &$61.20$        &$67.50$      &$55.22$        &$63.69$    &$51.48$       &$61.13$ \\
    \textbf{DFSL-DN4-KLD (ours)}    &with    &$\bm{75.35}$    &$\bm{63.92}$   &$\bm{69.16}$ &$\bm{60.26}$   &$\bm{66.96}$  &$\bm{57.55}$   &$\bm{65.25}$\\
    \textbf{DFSL-DN4-TCD (ours)}    &with    &$74.47$         &$62.35$        &$67.87$      &$58.53$        &$65.54$       &$\bm{55.79}$   &$63.79$\\
    \textbf{DFSL-DN4-SKL (ours)}    &with    &$74.85$         &$\bm{62.92}$   &$\bm{68.36}$ &$\bm{58.65}$   &$\bm{65.76}$  &$55.73$        &$\bm{63.89}$ \\
    \bottomrule
	\end{tabular}
	\label{comparison_FGSM_result_dog}
\end{table*}

\begin{table*}[!tp]
	\centering
    \extrarowheight=-1pt
	\caption{Comparison ($\%$) of different DFSL models on Stanford Cars under the $5$-way $5$-shot setting. Both training and test are based on a FGSM attacker. For each evaluation criterion, the best and the second best results are highlighted in bold.}
	\vspace{-3mm}
	\begin{tabular}{@{}lc|c|cc|cc|cc@{}}
	\toprule
    \multirow{2}{*}{Method} &\multirow{2}{*}{ET} &\multirow{2}{*}{$\mathcal{ACC}_\text{clean}$}   &\multicolumn{2}{c|}{$\epsilon\!=\!0.003$}  &\multicolumn{2}{c|}{$\epsilon\!=\!0.007$}  &\multicolumn{2}{c}{$\epsilon\!=\!0.01$} \\
                           & &  &$\mathcal{ACC}_\text{adv}$ &$\mathcal{F}_1$  &$\mathcal{ACC}_\text{adv}$ &$\mathcal{F}_1$ &$\mathcal{ACC}_\text{adv}$ &$\mathcal{F}_1$\\
    \midrule
    DN4+Ep                          &without  &$\bm{90.73}$   &$47.83$       &$62.63$       &$33.18$       &$48.59$      &$30.74$        &$45.92$ \\
    DFSL-DN4-AT                     &with     &$89.66$        &$75.30$       &$81.85$       &$63.33$       &$74.22$      &$56.08$        &$69.00$ \\
    DFSL-DN4-KL                     &with     &$85.71$        &$76.56$       &$80.87$       &$66.22$       &$74.71$      &$59.12$        &$69.97$ \\
    DFSL-DN4-ALP                    &with     &$87.60$        &$\bm{79.03}$  &$\bm{83.09}$  &$\bm{68.74}$  &$\bm{77.03}$ &$\bm{61.54}$   &$\bm{72.29}$ \\
    DFSL-DN4-Local-UDA              &with     &$87.55$        &$72.29$       &$79.19$       &$56.17$       &$68.43$      &$47.35$        &$61.46$ \\
    \textbf{DFSL-DN4-KLD (ours)}    &with     &$\bm{89.72}$   &$77.42$       &$\bm{83.11}$  &$63.44$       &$74.32$      &$54.94$        &$68.14$ \\
    \textbf{DFSL-DN4-TCD (ours)}    &with     &$89.33$        &$76.67$       &$82.51$       &$62.92$       &$73.83$      &$54.28$        &$67.52$ \\
    \textbf{DFSL-DN4-SKL (ours)}    &with     &$85.17$        &$\bm{79.09}$  &$82.01$       &$\bm{70.90}$  &$\bm{77.38}$ &$\bm{65.22}$   &$\bm{73.87}$ \\
    \bottomrule
	\end{tabular}
	\label{comparison_FGSM_result_car}
\end{table*}

\begin{table*}[!tp]
	\centering
    \extrarowheight=-1pt
	\caption{Comparison ($\%$) of different DFSL models on CUB Birds under the $5$-way $5$-shot setting. Both training and test are based on a FGSM attacker. For each evaluation criterion, the best and the second best results are highlighted in bold.}
	\vspace{-3mm}
	\begin{tabular}{@{}lc|c|cc|cc|cc@{}}
	\toprule
    \multirow{2}{*}{Method} &\multirow{2}{*}{ET} &\multirow{2}{*}{$\mathcal{ACC}_\text{clean}$}   &\multicolumn{2}{c|}{$\epsilon\!=\!0.003$}  &\multicolumn{2}{c|}{$\epsilon\!=\!0.007$}  &\multicolumn{2}{c}{$\epsilon\!=\!0.01$} \\
                           & &  &$\mathcal{ACC}_\text{adv}$ &$\mathcal{F}_1$  &$\mathcal{ACC}_\text{adv}$ &$\mathcal{F}_1$ &$\mathcal{ACC}_\text{adv}$ &$\mathcal{F}_1$\\
    \midrule
    DN4+Ep                         &without &$\bm{88.24}$  &$43.95$   &$58.67$   &$30.19$   &$44.98$    &$26.75$    &$41.05$ \\
    DFSL-DN4-AT                    &with    &$87.43$       &$75.46$   &$81.00$   &$69.56$   &$77.47$    &$66.15$    &$75.31$ \\
    DFSL-DN4-KL                    &with    &$87.69$       &$80.14$   &$83.74$   &$76.36$   &$81.63$    &$73.66$    &$80.06$ \\
    DFSL-DN4-ALP                   &with    &$83.89$       &$75.12$   &$79.26$   &$64.46$   &$72.90$    &$57.61$    &$68.30$ \\
    DFSL-DN4-Local-UDA             &with    &$84.00$       &$72.24$   &$77.67$   &$68.69$   &$75.57$    &$65.61$    &$73.67$ \\
    \textbf{DFSL-DN4-KLD (ours)}   &with    &$87.65$       &$\bm{80.51}$  &$\bm{83.92}$      &$\bm{77.68}$  &$\bm{82.36}$   &$\bm{75.04}$   &$\bm{80.85}$ \\
    \textbf{DFSL-DN4-TCD (ours)}   &with    &$87.17$       &$78.52$       &$82.61$           &$74.86$       &$80.54$        &$72.22$        &$78.99$ \\
    \textbf{DFSL-DN4-SKL (ours)}   &with    &$\bm{88.03}$  &$\bm{82.41}$  &$\bm{85.12}$      &$\bm{79.50}$  &$\bm{83.54}$   &$\bm{76.99}$   &$\bm{82.14}$ \\
    \bottomrule
	\end{tabular}
	\label{comparison_FGSM_result_bird}
\end{table*}

\subsection{Performing DFSL on Fine-grained Datasets}
\label{Section_5_4}
To show the consistent effectiveness of the proposed DFSL framework, we conduct experiments on three fine-grained image datasets, \textit{i.e.,} Stanford Dogs~\cite{khosla2011novel}, Stanford Cars~\cite{krause20133d} and CUB Birds-200-2011~\cite{wah2011caltech}, and report the results in Table~\ref{comparison_FGSM_result_dog}, Table~\ref{comparison_FGSM_result_car} and Table~\ref{comparison_FGSM_result_bird}, respectively. Because the data sizes of these three datasets are relatively small, we perform simple data augmentation on these datasets. All the other settings are similar to the settings in Section~\ref{Section_5_3}.

Specifically, we can see that DFSL-DN4-SKL (ours), as a prediction-wise consistency based method, can obtain consistently better results than other competitors on both clean accuracy and adversarial accuracy on all three fine-grained datasets. For example, on the CUB Birds-200-2011 dataset in Table~\ref{comparison_FGSM_result_bird}, when $\epsilon\!=\!0.01$, DFSL-DN4-SKL (ours) gains significantly improvements over DFSL-DN4-AT, DFSL-DN4-KL, DFSL-DN4-ALP and DFSL-DN4-Local-UDA by $10.84\%$, $3.33\%$, $19.38\%$ and $11.38\%$, respectively. Similarly, the proposed DFSL-DN4-TCD (ours) and  DFSL-DN4-KLD (ours) also perform superiorly on the three datasets. For example, DFSL-DN4-KLD (ours) achieves the best adversarial accuracies on all three levels of attacks on the Stanford Dogs dataset. Also, on the Stanford Cars dataset, DFSL-DN4-KLD (ours) obtains the best clean accuracy ($89.72\%$) than other adversarial defense methods, which is competitive to the result ($90.73\%$) of DN4+Ep, which is trained only using clean examples without any adversarial examples.
Typically, on both Stanford Dogs and CUB Birds-200-2011 datasets, our proposed DFSL methods (\textit{i.e.,} KLD, TCD and SKL) can achieve the highest or second highest $\mathcal{F}_1$ scores on all three levels of adversarial attacks.

In addition, we also conduct experiments with the 5-way 1-shot setting on the three fine-grained image datasets, where the results are reported in Tables~\ref{comparison_FGSM_result_dog2},~\ref{comparison_FGSM_result_car2} and~\ref{comparison_FGSM_result_bird2}, respectively. As seen, on all the three datasets, the proposed DFSL-DN4-SKL (ours) can obtain the best adversarial accuracy and the highest $\mathcal{F}_{1}$ score at all three levels of adversarial attacks. It is worth noting that our modified DFSL-DN4-Local-UDA also shows competitive results on the Stanford Cars dataset. This is because we use the richer local descriptors instead of the original global logit representations in~\cite{MMDICLR2019} for the UDA loss, which is more suitable for the DFSL setting.


\begin{table*}[!tp]
	\centering
    \extrarowheight=-1pt
	\caption{\textcolor{black}{Comparison ($\%$) of different DFSL models on Stanford Dogs under the $5$-way $1$-shot setting. Both training and test are based on a FGSM attacker. For each evaluation criterion, the best and the second best results are highlighted in bold.}}
	\vspace{-3mm}
	{\color{black}\begin{tabular}{@{}lc|c|cc|cc|cc@{}}
	\toprule
    \multirow{2}{*}{Method} &\multirow{2}{*}{ET} &\multirow{2}{*}{$\mathcal{ACC}_\text{clean}$}   &\multicolumn{2}{c|}{$\epsilon\!=\!0.003$}  &\multicolumn{2}{c|}{$\epsilon\!=\!0.007$}  &\multicolumn{2}{c}{$\epsilon\!=\!0.01$} \\
                           & &  &$\mathcal{ACC}_\text{adv}$ &$\mathcal{F}_1$  &$\mathcal{ACC}_\text{adv}$ &$\mathcal{F}_1$ &$\mathcal{ACC}_\text{adv}$ &$\mathcal{F}_1$\\
    \midrule
    DN4+Ep                          &without &$\bm{58.06}$    &$11.15$        &$18.70$       &$3.39$         &$6.40$        &$1.84$         &$3.56$ \\
    DFSL-DN4-AT                     &with    &$51.52$         &$\bm{43.01}$   &$46.88$       &$40.90$        &$45.59$       &$\bm{38.54}$   &$44.09$ \\
    DFSL-DN4-KL                     &with    &$51.21$         &$42.21$        &$46.27$       &$40.70$        &$45.35$       &$38.45$        &$43.92$ \\
    DFSL-DN4-ALP                    &with    &$51.52$         &$42.52$        &$46.58$       &$39.33$        &$44.60$       &$36.73$        &$42.88$ \\
    DFSL-DN4-Local-UDA              &with    &$51.02$         &$42.30$        &$46.25$       &$40.69$        &$45.27$       &$38.08$        &$43.61$ \\
    \textbf{DFSL-DN4-KLD (ours)}    &with    &$\bm{52.11}$    &$42.86$        &$\bm{47.03}$  &$\bm{40.98}$   &$\bm{45.87}$  &$38.51$        &$\bm{44.28}$\\
    \textbf{DFSL-DN4-TCD (ours)}    &with    &$50.65$         &$40.78$        &$45.18$       &$39.79$        &$44.56$       &$37.44$        &$43.05$\\
    \textbf{DFSL-DN4-SKL (ours)}    &with    &$50.82$         &$\bm{43.91}$   &$\bm{47.11}$  &$\bm{42.97}$   &$\bm{46.56}$  &$\bm{41.16}$   &$\bm{45.48}$ \\
    \bottomrule
	\end{tabular}}
	\label{comparison_FGSM_result_dog2}
	\vspace{-3mm}
\end{table*}

\begin{table*}[!tp]
	\centering
    \extrarowheight=-1pt
	\caption{\textcolor{black}{Comparison ($\%$) of different DFSL models on Stanford Cars under the $5$-way $1$-shot setting. Both training and test are based on a FGSM attacker. For each evaluation criterion, the best and the second best results are highlighted in bold.}}
	\vspace{-3mm}
	{\color{black}\begin{tabular}{@{}lc|c|cc|cc|cc@{}}
	\toprule
    \multirow{2}{*}{Method} &\multirow{2}{*}{ET} &\multirow{2}{*}{$\mathcal{ACC}_\text{clean}$}   &\multicolumn{2}{c|}{$\epsilon\!=\!0.003$}  &\multicolumn{2}{c|}{$\epsilon\!=\!0.007$}  &\multicolumn{2}{c}{$\epsilon\!=\!0.01$} \\
                           & &  &$\mathcal{ACC}_\text{adv}$ &$\mathcal{F}_1$  &$\mathcal{ACC}_\text{adv}$ &$\mathcal{F}_1$ &$\mathcal{ACC}_\text{adv}$ &$\mathcal{F}_1$\\
    \midrule
    DN4+Ep                          &without  &$\bm{59.65}$   &$19.23$       &$29.08$       &$8.65$        &$15.10$       &$5.68$         &$10.37$ \\
    DFSL-DN4-AT                     &with     &$47.41$        &$36.75$       &$41.40$       &$36.26$       &$41.09$       &$34.56$        &$39.97$ \\
    DFSL-DN4-KL                     &with     &$52.12$        &$40.68$       &$45.69$       &$38.03$       &$43.97$       &$35.97$        &$42.56$ \\
    DFSL-DN4-ALP                    &with     &$48.11$        &$37.43$       &$42.10$       &$35.44$       &$40.81$       &$33.77$        &$39.68$ \\
    DFSL-DN4-Local-UDA              &with     &$\bm{53.24}$   &$\bm{40.88}$  &$\bm{46.24}$  &$38.14$       &$\bm{44.44}$  &$36.22$        &$\bm{43.11}$ \\
    \textbf{DFSL-DN4-KLD (ours)}    &with     &$52.07$        &$38.99$       &$44.59$       &$\bm{38.34}$  &$44.16$       &$\bm{36.66}$   &$43.02$ \\
    \textbf{DFSL-DN4-TCD (ours)}    &with     &$50.25$        &$36.64$       &$42.37$       &$36.12$       &$42.02$       &$34.75$        &$41.08$ \\
    \textbf{DFSL-DN4-SKL (ours)}    &with     &$52.97$        &$\bm{41.78}$  &$\bm{46.71}$  &$\bm{40.18}$  &$\bm{45.69}$  &$\bm{38.50}$   &$\bm{44.59}$ \\
    \bottomrule
	\end{tabular}}
	\label{comparison_FGSM_result_car2}
	\vspace{-3mm}
\end{table*}

\begin{table*}[!tp]
	\centering
    \extrarowheight=-1pt
	\caption{\textcolor{black}{Comparison ($\%$) of different DFSL models on CUB Birds under the $5$-way $1$-shot setting. Both training and test are based on a FGSM attacker. For each evaluation criterion, the best and the second best results are highlighted in bold.}}
	\vspace{-3mm}
	{\color{black}\begin{tabular}{@{}lc|c|cc|cc|cc@{}}
	\toprule
    \multirow{2}{*}{Method} &\multirow{2}{*}{ET} &\multirow{2}{*}{$\mathcal{ACC}_\text{clean}$}   &\multicolumn{2}{c|}{$\epsilon\!=\!0.003$}  &\multicolumn{2}{c|}{$\epsilon\!=\!0.007$}  &\multicolumn{2}{c}{$\epsilon\!=\!0.01$} \\
                           & &  &$\mathcal{ACC}_\text{adv}$ &$\mathcal{F}_1$  &$\mathcal{ACC}_\text{adv}$ &$\mathcal{F}_1$ &$\mathcal{ACC}_\text{adv}$ &$\mathcal{F}_1$\\
    \midrule
    DN4+Ep                         &without &$\bm{71.92}$  &$28.89$       &$41.22$       &$14.40$       &$23.99$      &$10.42$        &$18.20$ \\
    DFSL-DN4-AT                    &with    &$67.08$       &$54.37$       &$60.05$       &$42.24$       &$51.83$      &$36.28$        &$47.09$ \\
    DFSL-DN4-KL                    &with    &$\bm{69.36}$  &$54.61$       &$61.10$       &$41.41$       &$51.85$      &$35.14$        &$46.64$ \\
    DFSL-DN4-ALP                   &with    &$66.30$       &$\bm{58.23}$  &$\bm{62.00}$  &$\bm{47.80}$  &$\bm{55.55}$ &$\bm{42.25}$   &$\bm{51.61}$ \\
    DFSL-DN4-Local-UDA             &with    &$66.40$       &$54.24$       &$59.70$       &$43.22$       &$52.35$      &$37.73$        &$48.11$ \\
    \textbf{DFSL-DN4-KLD (ours)}   &with    &$66.75$       &$55.21$       &$60.43$       &$43.74$       &$52.84$      &$37.95$        &$48.38$ \\
    \textbf{DFSL-DN4-TCD (ours)}   &with    &$66.75$       &$55.37$       &$60.52$       &$44.24$       &$53.21$      &$38.58$        &$48.89$ \\
    \textbf{DFSL-DN4-SKL (ours)}   &with    &$66.78$       &$\bm{58.63}$  &$\bm{62.44}$  &$\bm{48.48}$  &$\bm{56.17}$ &$\bm{43.09}$   &$\bm{52.38}$ \\
    \bottomrule
	\end{tabular}}
	\label{comparison_FGSM_result_bird2}
	\vspace{-3mm}
\end{table*}

\subsection{Performing DFSL with PGD-adversarial Training}
\label{Section_5_8}
Besides the FGSM attacker, we also apply a PGD~\cite{madry2017towards} attacker to train more robust DFSL models for defending against various stronger attacks. Specifically, we set the fixed perturbation $\epsilon$, the number of iteration steps and step size of PGD as $0.02$, $10$ and $1/255$, respectively, \emph{i.e.,} a PGD-10 attacker is used to adversarially train all the DFSL models. Moreover, various attackers, such as PGD-10~\cite{madry2017towards}, DeepFool~\cite{Deepfool_CVPR_2016}, C$\&$W~\cite{CW2017} and FGSM~\cite{goodfellow2014explaining}, are adopted to verify the robustness of PGD-adversarial trained DFSL models. For the DeepFool, the maximum number of iteration is set as $30$. As for C$\&$W, the learning rate and the maximum iteration number are set as $0.001$ and $10$, respectively.

The results on \emph{mini}ImageNet under both 5-shot and 1-shot settings are reported in Tables~\ref{comparison_PGD_result_mini} and~\ref{comparison_PGD_result_mini2}. Comparing DFSL-DN4-AT (PGD) with DFSL-DN4-AT (FGSM) in Table~\ref{comparison_PGD_result_mini}, we can see that although the FGSM-adversarial trained model enjoys a great clean accuracy and has a good defense ability for the FGSM attack, it fails to defend against the stronger attacks, \emph{e.g.,} PGD, DeepFool and C$\&$W. In contrast, the PGD-adversarial trained model, \emph{i.e.,} DFSL-DN4-AT (PGD), has a much stronger defense ability for these stronger attacks, which gains more than $10\%$ adversarial accuracy improvements under the DeepFool and C$\&$W attacks, and gains more than $20\%$ adversarial accuracy improvements under the PGD attack over DFSL-DN4-AT (FGSM). Nevertheless, the standard AT will suffer from a loss on the clean accuracy with a drop of $4.9\%$.

On the contrary, the proposed KLD, TCD and SKL can not only improve the clean accuracy, but also significantly improves the adversarial accuracy under all kinds of attacks. For example, TCD obtains $8.52\%$, $1.56\%$, $2.76\%$ and $4.61\%$ adversarial accuracy improvements over AT (PGD) under the attacks of PGD, DeepFool, C$\&$W and FGSM, respectively. The similar results can also be observed in Table~\ref{comparison_PGD_result_mini2} under the 5-way 1-shot setting. The above results indicate that the proposed methods could consistently gain improvements over AT (PGD) in terms of both the clean accuracy and adversarial accuracy.

\begin{table*}[!tp] 
	\centering
    \extrarowheight=-1pt
	\caption{\textcolor{black}{Comparison ($\%$) of different DFSL models with PGD-adversarial training on \textit{mini}ImageNet under the $5$-way $5$-shot setting. For each evaluation criterion, the best and the second best results are highlighted in bold.}}
	\vspace{-3mm}
	\begin{tabular}{@{}lcc|c|cc|cc|cc|cc@{}}
	\toprule
    \multirow{2}{*}{Method} &\multirow{2}{*}{ET} &\multirow{2}{*}{Defense} &\multirow{2}{*}{$\mathcal{ACC}_\text{clean}$}   &\multicolumn{2}{c|}{PGD-10}  &\multicolumn{2}{c|}{DeepFool}  &\multicolumn{2}{c|}{C\&W}    &\multicolumn{2}{c}{FGSM ($\epsilon\!=\!0.01$)}\\
                           & & &  &$\mathcal{ACC}_\text{adv}$ &$\mathcal{F}_1$  &$\mathcal{ACC}_\text{adv}$ &$\mathcal{F}_1$ &$\mathcal{ACC}_\text{adv}$ &$\mathcal{F}_1$ &$\mathcal{ACC}_\text{adv}$ &$\mathcal{F}_1$\\
    \midrule
    DFSL-DN4-AT                   &with &FGSM &$\bm{71.86}$  &$28.60$      &$40.91$   &$37.25$      &$49.06$      &$19.57$       &$30.76$         &$58.51$   &$64.50$\\
    DFSL-DN4-AT                   &with &PGD  &$66.96$       &$50.31$      &$57.45$       &$48.99$      &$56.58$      &$29.63$       &$41.08$      &$59.86$        &$63.21$\\
    \textbf{DFSL-DN4-KLD (ours)}  &with &PGD  &$68.68$       &$\bm{58.03}$ &$62.90$       &$\bm{50.67}$ &$\bm{58.31}$ &$\bm{31.90}$  &$\bm{43.56}$  &$\bm{64.58}$   &$\bm{66.56}$\\
    \textbf{DFSL-DN4-TCD (ours)}  &with &PGD  &$68.43$       &$\bm{58.83}$ &$\bm{63.26}$  &$\bm{50.55}$ &$\bm{58.14}$ &$\bm{32.39}$  &$\bm{43.96}$  &$\bm{64.47}$   &$66.39$\\
    \textbf{DFSL-DN4-SKL (ours)}  &with &PGD  &$\bm{68.98}$  &$58.04$      &$\bm{63.03}$  &$42.21$      &$52.37$      &$22.77$       &$34.23$      &$64.48$         &$\bm{66.65}$ \\
    \bottomrule
	\end{tabular}
	\label{comparison_PGD_result_mini}
	\vspace{-3mm}
\end{table*}

\begin{table*}[!tp] 
	\centering
    \extrarowheight=-1pt
	\caption{\textcolor{black}{Comparison ($\%$) of different DFSL models with PGD-adversarial training on \textit{mini}ImageNet under the $5$-way $1$-shot setting. For each evaluation criterion, the best and the second best results are highlighted in bold.}}
	\vspace{-3mm}
	\begin{tabular}{@{}lcc|c|cc|cc|cc|cc@{}}
	\toprule
    \multirow{2}{*}{Method} &\multirow{2}{*}{ET} &\multirow{2}{*}{Defense} &\multirow{2}{*}{$\mathcal{ACC}_\text{clean}$}   &\multicolumn{2}{c|}{PGD-10}  &\multicolumn{2}{c|}{DeepFool}  &\multicolumn{2}{c|}{C\&W}    &\multicolumn{2}{c}{FGSM ($\epsilon\!=\!0.01$)}\\
                           & & &  &$\mathcal{ACC}_\text{adv}$ &$\mathcal{F}_1$  &$\mathcal{ACC}_\text{adv}$ &$\mathcal{F}_1$ &$\mathcal{ACC}_\text{adv}$ &$\mathcal{F}_1$ &$\mathcal{ACC}_\text{adv}$ &$\mathcal{F}_1$\\
    \midrule
    DFSL-DN4-AT                   &with &FGSM  &$\bm{50.00}$  &$26.40$    &$34.55$   &$35.73$      &$41.67$      &$10.52$       &$17.38$    &$40.77$   &$44.91$\\
    DFSL-DN4-AT                   &with &PGD   &$47.87$       &$37.72$    &$42.19$   &$38.09$      &$42.42$      &$16.06$       &$24.05$    &$43.55$   &$45.60$\\
    \textbf{DFSL-DN4-KLD (ours)} &with &PGD  &$47.93$      &$37.35$       &$41.98$       &$\bm{39.24}$   &$\bm{43.15}$   &$17.01$        &$25.10$       &$43.89$       &$45.82$\\
    \textbf{DFSL-DN4-TCD (ours)} &with &PGD  &$48.34$      &$\bm{39.11}$  &$\bm{43.23}$  &$\bm{39.44}$   &$\bm{43.43}$   &$\bm{17.41}$   &$\bm{25.60}$  &$\bm{44.55}$  &$\bm{46.36}$\\
    \textbf{DFSL-DN4-SKL (ours)} &with &PGD  &$\bm{48.79}$ &$\bm{40.17}$  &$\bm{44.06}$  &$38.43$        &$42.99$        &$\bm{17.94}$   &$\bm{26.23}$  &$\bm{45.01}$  &$\bm{46.82}$ \\
    \bottomrule
	\end{tabular}
	\label{comparison_PGD_result_mini2}
	\vspace{-3mm}
\end{table*}

\begin{table*}[!tp]
	\centering
    \extrarowheight=-1pt
	\caption{Cross-domain defense transferability of DFSL when the attack level is $\epsilon\!=\!0.01$. Both training and test are based on a FGSM attacker under the $5$-way $5$-shot setting. For each evaluation criterion, the best and the second best results are highlighted in bold.}
	\vspace{-3mm}
	\begin{tabular}{@{}lc|ccc|ccc|ccc@{}}
	\toprule
    \multirow{2}{*}{Method} &\multirow{2}{*}{ET}  &\multicolumn{3}{c|}{\emph{mini}ImageNet$\rightarrow$Stanford Dogs}  &\multicolumn{3}{c|}{\emph{mini}ImageNet$\rightarrow$Stanford Cars}  &\multicolumn{3}{c}{\emph{mini}ImageNet$\rightarrow$CUB Birds} \\
                            &  &$\mathcal{ACC}_\text{clean}$  &$\mathcal{ACC}_\text{adv}$ &$\mathcal{F}_1$  &$\mathcal{ACC}_\text{clean}$ &$\mathcal{ACC}_\text{adv}$ &$\mathcal{F}_1$ &$\mathcal{ACC}_\text{clean}$ &$\mathcal{ACC}_\text{adv}$ &$\mathcal{F}_1$\\
    \midrule
    DN4+AT                          &without  &$37.30$       &$22.31$        &$27.92$       &$38.60$       &$24.18$       &$29.73$       &$46.88$       &$33.87$       &$39.32$ \\
    DFSL-DN4-AT                     &with     &$55.89$       &$38.67$        &$45.71$       &$48.71$       &$34.54$       &$40.41$       &$65.23$       &$44.21$       &$52.70$\\
    \textbf{DFSL-DN4-KLD (ours)}    &with     &$\bm{56.30}$  &$42.01$        &$48.11$       &$\bm{49.73}$  &$37.69$       &$42.88$       &$\bm{66.20}$  &$52.95$       &$58.83$\\
    \textbf{DFSL-DN4-TCD (ours)}    &with     &$\bm{56.52}$  &$\bm{42.94}$   &$\bm{48.80}$  &$\bm{50.31}$  &$\bm{39.27}$  &$\bm{44.10}$  &$\bm{65.81}$  &$\bm{53.75}$  &$\bm{59.17}$\\
    \textbf{DFSL-DN4-SKL (ours)}    &with     &$55.23$       &$\bm{44.83}$   &$\bm{49.48}$  &$49.05$       &$\bm{40.68}$  &$\bm{44.47}$  &$65.31$       &$\bm{57.01}$  &$\bm{60.87}$\\
    \bottomrule
	\end{tabular}
	\label{comparison_cross_domain_result}
	\vspace{-3mm}
\end{table*}

\subsection{Performing DFSL on Cross-domain Datasets}
\label{Section_5_5}
It will be interesting to further investigate the defense transfer ability of the proposed DFSL framework in cross-domain scenarios. To this end, following the cross-domain FSL work in the literature~\cite{chen2019closer}, we conduct an experiment on three cross-domain scenarios, \textit{i.e.,} \emph{mini}ImageNet$\rightarrow$Stanford Dogs, \emph{mini}ImageNet$\rightarrow$Stanford Cars and \emph{mini}ImageNet$\rightarrow$CUB Birds. In this experiment, all the models are adversarially trained on the source domain dataset (\textit{i.e.,} \emph{mini}ImageNet) and directly tested on the target domain dataset (\textit{e.g.,} CUB Birds) without fine-tuning, by using a FGSM attacker with the attack level of $\epsilon\!=\!0.01$.

\begin{figure*}[!tbp]
\centering
\subfigure{
           \includegraphics[width=0.22\textwidth]{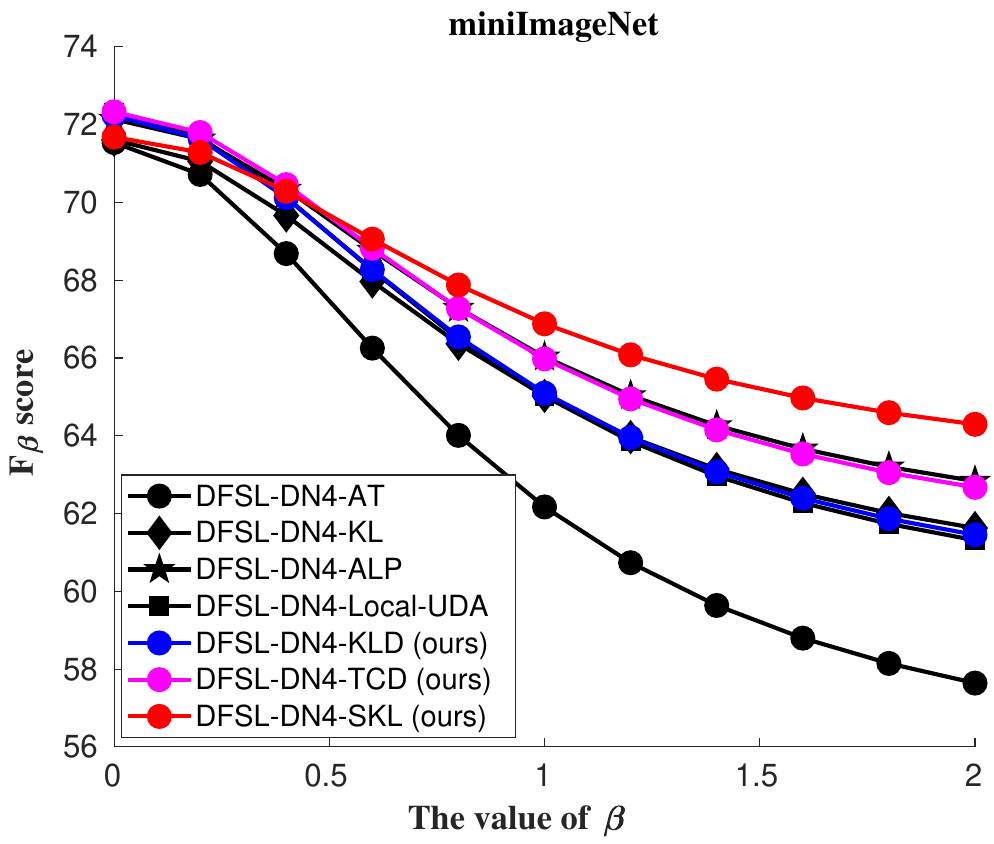}}
\subfigure{
           \includegraphics[width=0.22\textwidth]{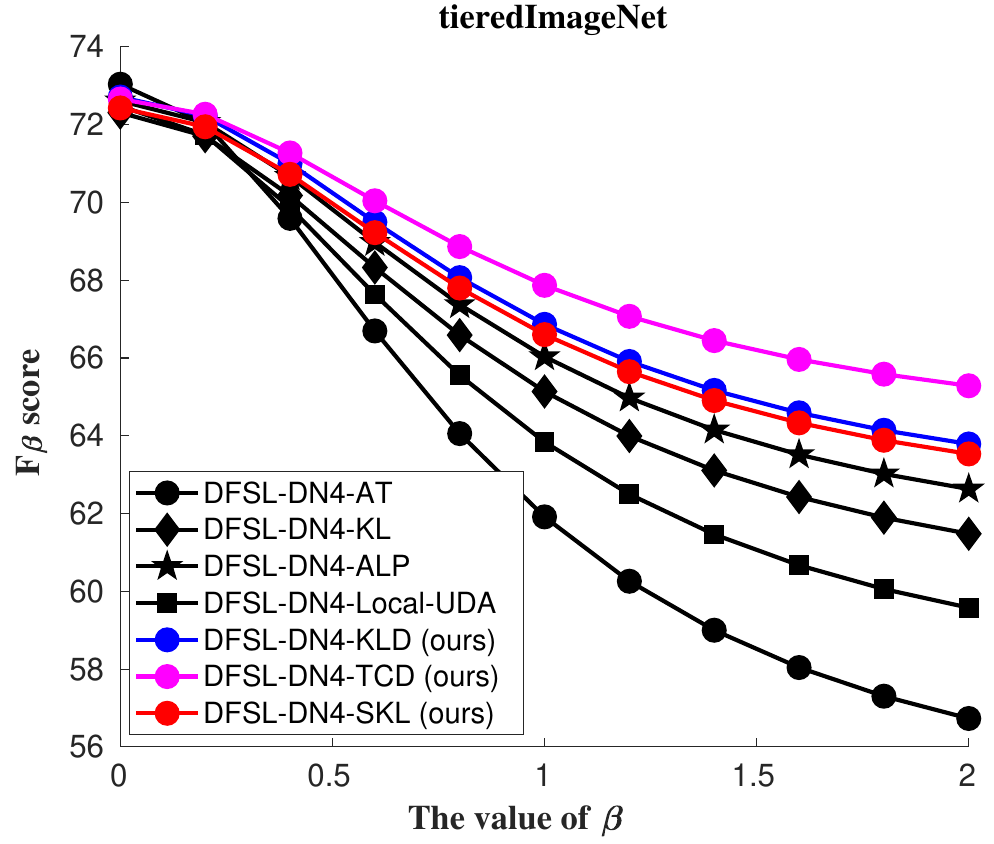}}
\subfigure{
           \includegraphics[width=0.22\textwidth]{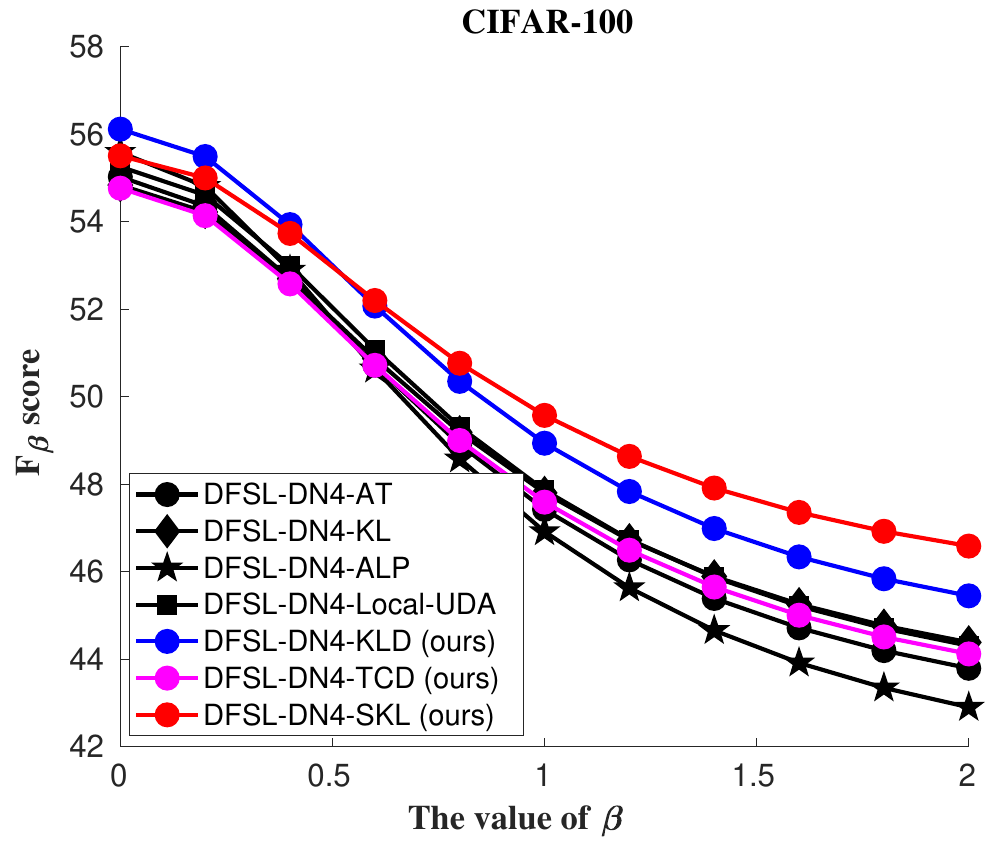}}\\
\subfigure{
           \includegraphics[width=0.22\textwidth]{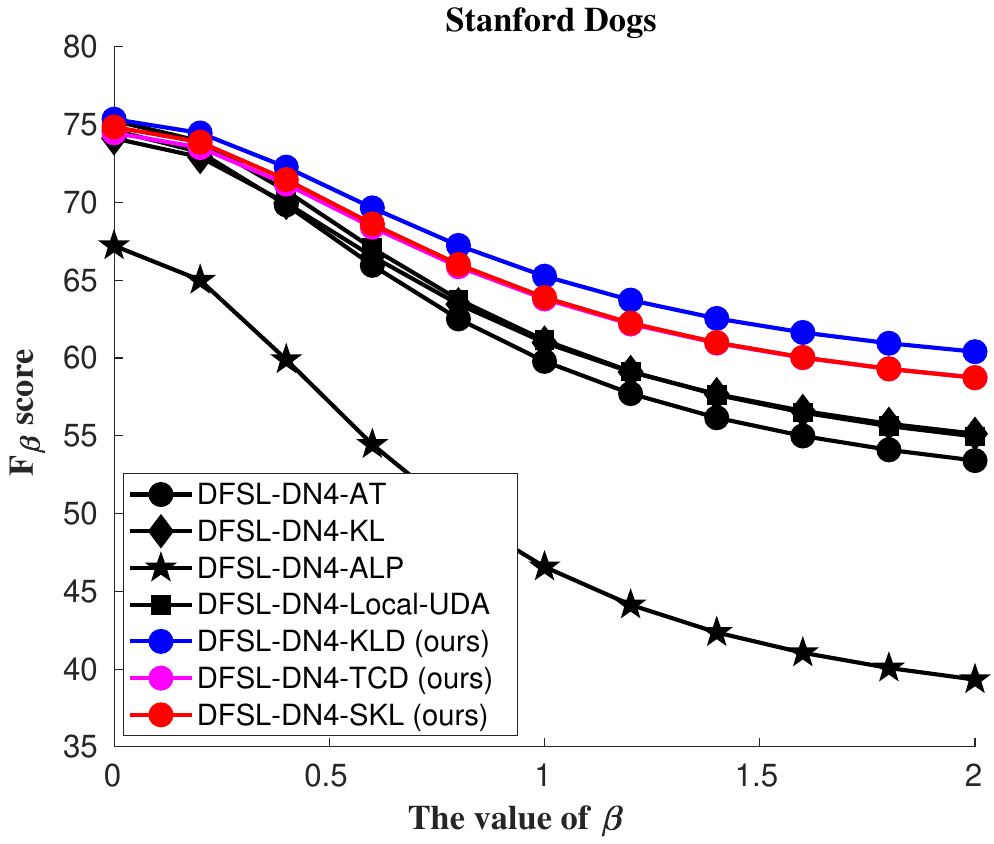}}
\subfigure{
           \includegraphics[width=0.22\textwidth]{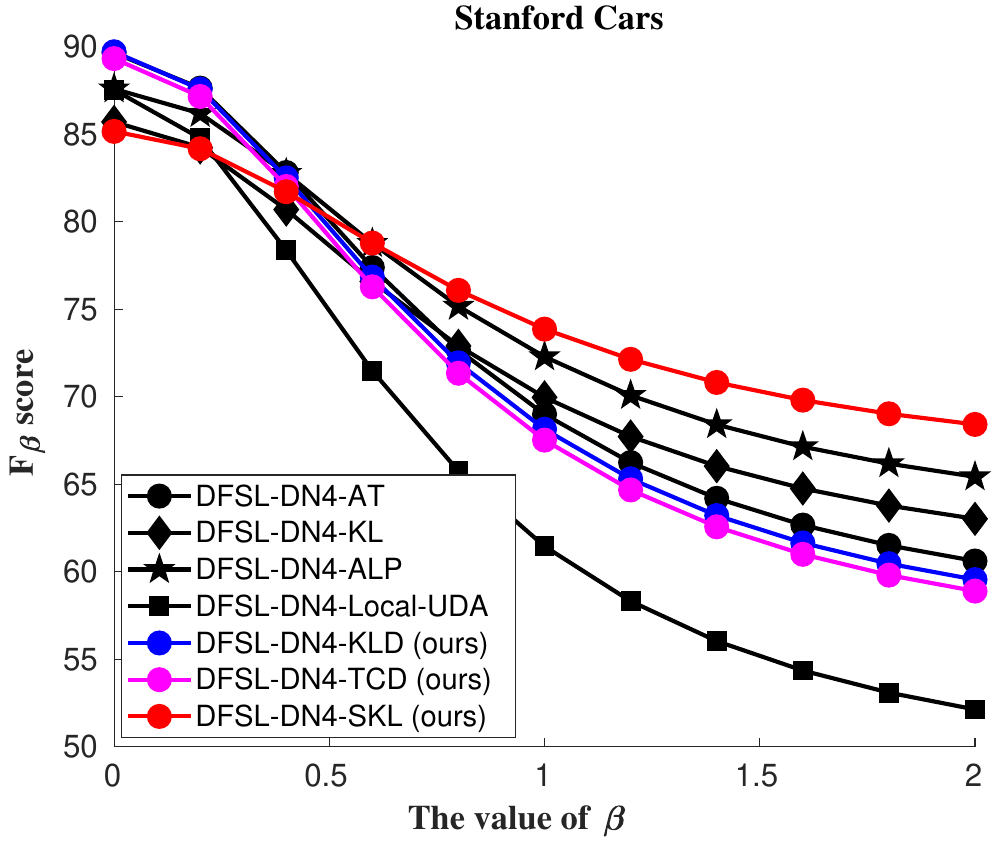}}
\subfigure{
           \includegraphics[width=0.22\textwidth]{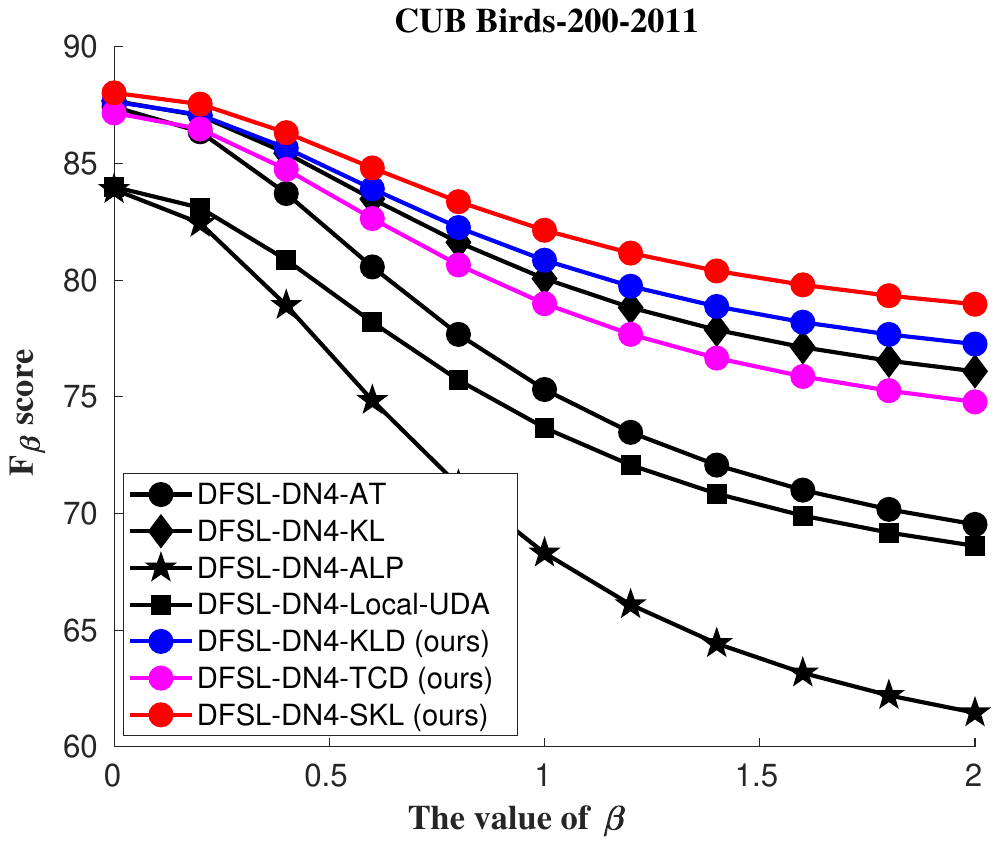}}
\vspace{-3mm}
\caption{$\mathcal{F}_{\beta}$ curves of different DFSL models on six datasets, by varying the value of $\beta$ from $0$ to $2$, where the attack level is $\epsilon=0.01$.}
\label{fig_F_beta_curve}
\end{figure*}

Specifically, DN4+AT, DFSL-DN4-AT, DFSL-DN4-KLD (ours), DFSL-DN4-TCD (ours) and DFSL-DN4-SKL (ours) are selected as representatives. From Table~\ref{comparison_cross_domain_result}, we can observe that DFSL-DN4-AT can significantly improve both the clean and adversarial accuracies over DN4+AT. It verifies that ET (\textit{i.e.,} Ep+AT) indeed has the ability of transferring defense knowledge even in cross-domain scenarios. More importantly, DFSL-DN4-KLD (ours), DFSL-DN4-TCD (ours) and DFSL-DN4-SKL (ours) can further improve the adversarial accuracy over DFSL-DN4-AT, which also further verifies that the proposed feature-wise or prediction-wise distribution consistency criteria are effective.

\begin{figure}[!tbp]
\centering
\subfigure{
           \includegraphics[width=0.2\textwidth]{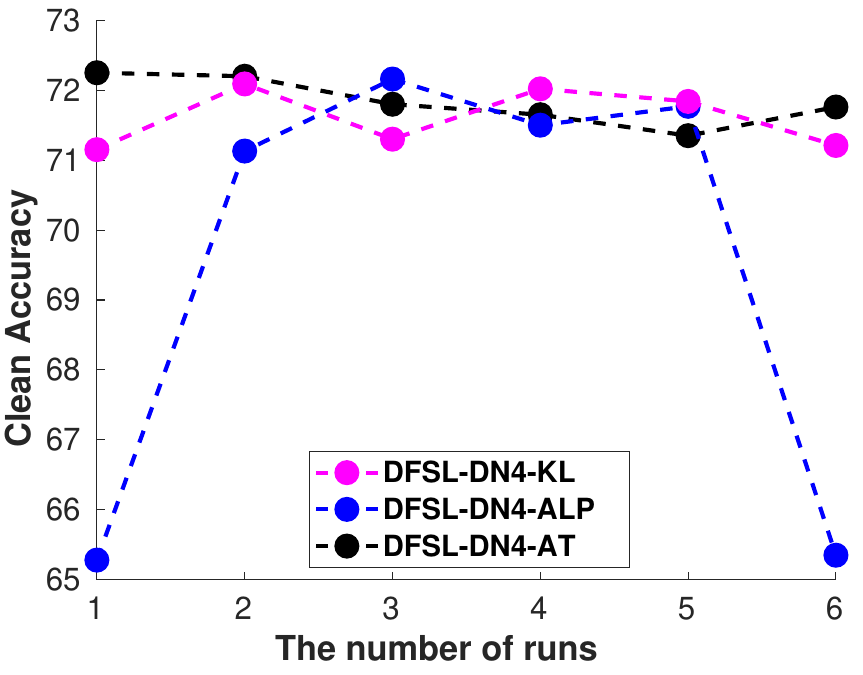}}
\subfigure{
           \includegraphics[width=0.2\textwidth]{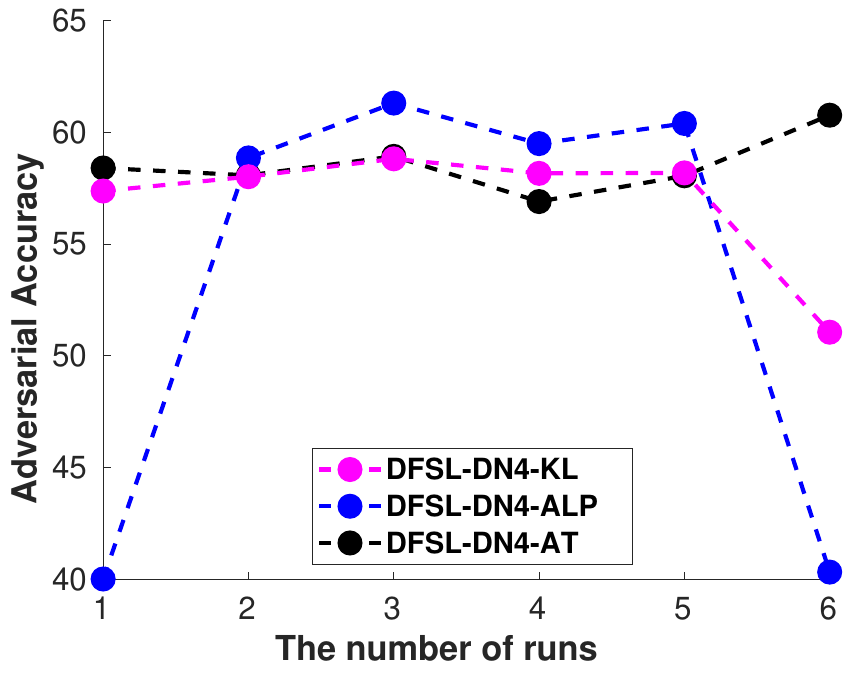}}
\vspace{-2mm}
\caption{Randomness matters. On the same server with the same codebase, DFSL-DN4-AT, DFSL-DN4-ALP and DFSL-DN4-KL are repeatedly run (trained) six times, where only the seed of generating random number is varied. Left subfigure: clean accuracy of these models at different runs; Right subfigure: adversarial accuracy ($\epsilon\!=\!0.01$) of these models at different runs.}
\label{fig_randomness}
\end{figure}

\subsection{Randomness Matters}
\label{Section_5_6}
There is a highly important but easily overlooked issue in both fields of the generic adversarial training and generic few-shot learning in the literature, that is the non-reproducibility of one model caused by the randomness. One clear consequence of ignoring the randomness is that the fairness of the comparison cannot be guaranteed. In other words, sometimes the improvement obtained may simply be due to the randomness.

Typically, we can roughly summarize the randomness into five categories: (1) randomness introduced by different hardware platforms (\textit{e.g.,} different servers); (2) randomness introduced by different software platforms (\textit{e.g.,} different deep learning platforms or releases); (3) randomness introduced by different codebases, including the optimizer, learning rate and training epochs, \textit{etc}; (4) randomness introduced by different initializations of network parameters and different data shuffles; (5) randomness introduced by the nondeterminism of CUDA and CuDNN backends.

To fix the first two kinds of randomness, we run all the comparison methods in the same server and use the same Pytorch release. For the third kind of randomness, we re-implement all the comparison methods with the same single codebase including using the same settings of optimization and learning rate, except the core parts of the method itself and hyper-parameters. As for the fourth kind of randomness, we set the same fixed seed for all comparison methods. For the last kind of randomness, we seed manually for CUDA and make CuDNN deterministic. In this way, we can make sure that both the initialization of network parameters and data shuffling is the same for all the comparison methods, and the results of any methods are reproducible on the same server.

To further demonstrate the randomness's impact on the final results, we fix the first three kinds of randomness and repeatedly run a method multiple times. Specifically, under the $5$-way $5$-shot setting, we run \textit{DFSL-DN4-AT}, \textit{DFSL-DN4-ALP} and \textit{DFSL-DN4-KL} six times on \emph{mini}ImageNet, respectively. The results are plotted in Fig.~\ref{fig_randomness}. As seen, although the computing environment and code haven't changed at all, the performance of each method varies significantly because of the random initializations (seeds) at different runs. In this sense, we cannot draw a conclusion that one method performs strictly better than another method. Therefore, in all of our experiments, we have well fixed the randomness to strictly compare different methods for fairness.

\subsection{Qualitative Comparison: Curve of $\mathcal{F}_{\beta}$ Scores}
\label{Section_5_7}
In addition to the $\mathcal{F}_1$ score, we can also generate the curves of $\mathcal{F}_{\beta}$ scores by varying the value of $\beta$ for qualitative comparison. To be specific, we first calculate a series of $F_{\beta}$ scores with the clean accuracy $\mathcal{ACC}_\text{clean}$ and adversarial accuracy $\mathcal{ACC}_\text{adv}$ according to Eq.~(\ref{fun11}) by varying $\beta$ from $0$ to $2$. After that, we can plot a $F_{\beta}$ curve via these $F_{\beta}$ scores for each DFSL model. With the results in Tables~\ref{comparison_FGSM_result_mini},~\ref{comparison_FGSM_result_tier},~\ref{comparison_FGSM_result_cifar},~\ref{comparison_FGSM_result_dog},~\ref{comparison_FGSM_result_car} and~\ref{comparison_FGSM_result_bird}, the $F_{\beta}$ curves are worked out and plotted in Fig.~\ref{fig_F_beta_curve}. As seen, in most cases, the proposed DFSL-DN4-KLD, DFSL-DN4-TCD and DFSL-DN4-SKL perform consistently superior to other competitors for any value of $\beta$.

\section{Conclusions}
In this paper, we propose a new challenging issue for the first time, \textit{i.e.}, \textit{defensive few-shot learning (DFSL)}, aiming to learn robust few-shot models against adversarial attacks. To tackle this issue, we propose a unified DFSL framework with solutions from two aspects, \textit{i.e.}, task-level distribution consistency and distribution consistency within each task. Extensive experiments have verified that: (1) the proposed \textit{episode-based adversarial training (ET) mechanism} can effectively transfer adversarial defense knowledge by leveraging the task-level distribution consistency; (2) the proposed feature-wise and prediction-wise consistency criteria, \textit{i.e.,}  \textit{Kullback-Leibler divergence based distribution measure (KLD)}, \textit{task-conditioned distribution measure (TCD)}, and \textit{Symmetric Kullback-Leibler divergence measure (SKL)} can reliably narrow the distribution gap between the clean and adversarial examples and enjoy good generalization performance. Moreover, we modify and re-implement multiple existing adversarial defense methods and multiple representative FSL methods into this unified framework as well as rich baseline results, which can significantly facilitate future research on the topic of DFSL. In addition, we propose a unified evaluation criterion, \textit{i.e.,} $\mathcal{F}_{\beta}$ scores, which is also of significance for the community.
Many future directions are worth exploring for the new topic of DFSL. Especially, we are going to extend the proposed DFSL framework to the transductive setting and further investigate the cross-domain scenarios on the large-scale dataset, \emph{e.g.,} Meta-Dataset~\cite{Meta_dataset_ICLR2020}.

\section*{Acknowledgements}
This work is supported in part by the National Natural Science Foundation of China (62192783, 62106100, 62276128), Jiangsu Natural Science Foundation (BK20221441), CAAI-Huawei MindSpore Open Fund, the Collaborative Innovation Center of Novel Software Technology and Industrialization, and Jiangsu Provincial Double-Innovation Doctor Program (JSSCBS20210021).





\ifCLASSOPTIONcaptionsoff
  \newpage
\fi



%

\bibliographystyle{IEEEtran}
\bibliography{egbib}

\end{document}